\documentclass{article}

\usepackage[final]{nips_2017}

\usepackage[square,sort,comma,numbers]{natbib}

\usepackage[utf8]{inputenc} 
\usepackage[T1]{fontenc}    
\usepackage{hyperref}       
\usepackage{url}            
\usepackage{booktabs}       
\usepackage{amsfonts}       
\usepackage{nicefrac}       
\usepackage{microtype}      

\usepackage{graphicx}
\usepackage{subfigure} 
\usepackage{algorithm}
\usepackage{algorithmic}
\usepackage{mathtools}
\usepackage{bm}
\usepackage{color}
\usepackage{wrapfig}

\newtheorem{theorem}{Theorem}
\newtheorem{lemma}{Lemma}

\newtheorem{definition}{Definition}
\newtheorem{remark}{Remark}
\newenvironment{proof}{\paragraph{Proof:}}

\def\toptitlebar{\hrule height4pt\vskip .25in\vskip-\parskip}
\def\bottomtitlebar{\vskip .29in\vskip-\parskip\hrule height1pt\vskip .09in}

\title{VisualBackProp: efficient visualization of CNNs}

\author{
  Mariusz Bojarski \\
  NVIDIA Corp.\\
  \texttt{mbojarski@nvidia.com} \\
  \And
  Anna Choromanska \\
  ECE NYU Tandon\\
  \texttt{ac5455@nyu.edu} \\
  \And
  Krzysztof Choromanski \\
  Google Brain Robotics \\
  \texttt{kchoro@google.com} \\
  \And
  Bernhard Firner \\
  NVIDIA Corp.\\
  \texttt{bfirner@nvidia.com} \\
  \And
  Larry Jackel \\
  NVIDIA Corp.\\
  \texttt{ljackel@nvidia.com} \\
  \And
  Urs Muller \\
  NVIDIA Corp.\\
  \texttt{umuller@nvidia.com} \\
  \And
  Karol Zieba \\
  NVIDIA Corp.\\
  \texttt{kzieba@nvidia.coml} \\
}

\begin{document}

\maketitle

\begin{abstract}
This paper proposes a new method, that we call VisualBackProp, for visualizing which sets of pixels of the input image contribute most to the predictions made by the convolutional neural network (CNN). The method heavily hinges on exploring the intuition that the feature maps contain less and less irrelevant information to the prediction decision when moving deeper into the network. The technique we propose was developed as a debugging tool for CNN-based systems for steering self-driving cars and is therefore required to run in real-time, i.e. it was designed to require less computations than a forward propagation. This makes the presented visualization method a valuable debugging tool which can be easily used during both training and inference. We furthermore justify our approach with theoretical arguments and theoretically confirm that the proposed method identifies sets of input pixels, rather than individual pixels, that collaboratively contribute to the prediction. Our theoretical findings stand in agreement with the experimental results. The empirical evaluation shows the plausibility of the proposed approach on the road video data as well as in other applications and reveals that it compares favorably to the layer-wise relevance propagation approach, i.e. it obtains similar visualization results and simultaneously achieves order of magnitude speed-ups. 
\end{abstract}

\section{Introduction}
\label{sec:intro}

A plethora of important real-life problems are currently addressed with CNNs~\cite{LeCun:1989:BAH:1351079.1351090}, including image recognition~\cite{NIPS2012_4824}, speech recognition~\cite{DBLP:conf/icassp/Abdel-HamidMJP12}, and natural language processing~\cite{DBLP:conf/emnlp/WestonCA14}. More recently they were successfully used in the complex intelligent autonomous systems such as self-driving cars~\cite{DBLP:journals/corr/BojarskiTDFFGJM16, conf/iccv/ChenSKX15}. One of the fundamental question that arises when considering CNNs as well as other deep learning models is: \textit{what made the trained neural network model arrive at a particular response?} This question is of particular importance to the end-to-end systems, where the interpretability of the system is limited. Visualization tools aim at addressing this question by identifying parts of the input image that had the highest influence on forming the final prediction by the network. It is also straightforward to think about visualization methods as a debugging tool that helps to understand if the network detects ``reasonable'' cues from the image to arrive at a particular decision. 

\begin{figure}[!htp]
\centering
\vspace{-0.05in}
\includegraphics[width=0.245\columnwidth]{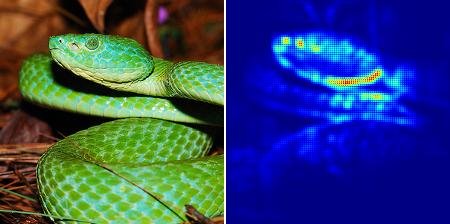}
\includegraphics[width=0.245\columnwidth]{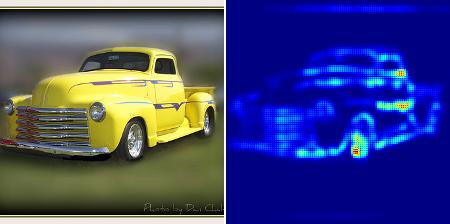} 
\includegraphics[width=0.245\columnwidth]{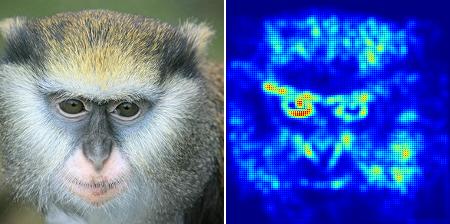} 
\vspace{-0.05in}
\caption{Visualization masks obtained by VisualBackProp on ImageNet for ResNet-$200$~\cite{He2016IdentityMI}.}
\vspace{-0.1in}
\label{fig:example}
\end{figure}

The visualization method for CNNs proposed in this paper was originally developed for CNN-based systems for steering autonomous cars, though it is highly general and can be used in other applications as well. The method relies on the intuition that when moving deeper into the network, the feature maps contain less and less information which are irrelevant to the output. Thus, the feature maps of the last convolutional layer should contain the most relevant information to determine the output. At the same time, feature maps of deeper layers have lower resolution. The underlying idea of the approach is to combine feature maps containing only relevant information (deep ones) with the ones with higher resolution (shallow ones). In order to do so, starting from the feature maps of the last convolutional layer, we ``backpropagate'' the information about the regions of relevance while simultaneously increasing the resolution, where the backpropagation procedure is not gradient-based (as is the case for example in sensitivity-based approaches~\cite{Baehrens:2010:EIC:1756006.1859912,DBLP:journals/corr/SimonyanVZ13,Rasmussen_visualizationof}), but instead is value-based. We call this approach VisualBackProp (an exemplary results are demonstrated in Figure~\ref{fig:example}).

Our method provides a general tool for verifying that the predictions generated by the neural network\footnote{In the case of our end-to-end system for steering an autonomous car, a prediction is the steering wheel angle.}, are based on reasonable optical cues in the input image. In case of autonomous driving these can be lane markings, other cars, or edges of the road. Our visualization tool runs in real time and requires less computations than forward propagation. We empirically demonstrate that it is order of magnitude faster than the state-of-the-art visualization method, layer-wise relevance propagation (LRP)~\cite{BachPLOS15}, while at the same time it leads to very similar visualization results.

In the theoretical part of this paper we first provide a rigorous mathematical analysis of the contribution of input neurons to the activations in the last layer that relies on network flows. We propose a quantitative measure of that contribution. We then show that our algorithm finds for each neuron the approximated value of that measure. To the best of our knowledge, the majority of the existing visualization techniques for deep learning, which we discuss in the Related Work section, lack theoretical guarantees, which instead we provide for our approach. This is yet another important contribution of this work.

The paper is organized as follows: Section~\ref{sec:rel_work} discusses related work, Section~\ref{sec:vism} describes our approach and Section~\ref{sec:theory} provides its theoretical analysis, Section~\ref{sec:exper} presents empirical results, and Section~\ref{sec:concl} contains final remarks.

\section{Related work}
\label{sec:rel_work}

A notable approach~\cite{BachPLOS15} addressing the problem of understanding classification decisions by pixel-wise decomposition of non-linear classifiers proposes a methodology called layer-wise relevance propagation, where the prediction is back-propagated without using gradients such that the relevance of each neuron is redistributed to its predecessors through a particular message-passing scheme relying on the conservation principle. The stability of the method and the sensitivity to different settings of the conservation parameters was studied in the context of several deep learning models~\cite{BinBacMonMueSam16}. The LRP technique was extended to Fisher Vector classifiers~\cite{bach-arxiv15} and also used to explain predictions of CNNs in NLP applications~\cite{DBLP:journals/corr/ArrasHMMS16}. An extensive comparison of LRP with other techniques, like the deconvolution method~\cite{MR14} and the sensitivity-based approach~\cite{DBLP:journals/corr/SimonyanVZ13}, which we also discuss next in this section, using an evaluation based on region perturbation can be found in~\cite{DBLP:journals/corr/SamekBMBM15}. This study reveals that LRP provides better explanation of the DNN classification decisions than considered competitors\footnote{We thus chose LRP as a competitive technique to our method in the experimental section.}.

Another approach~\cite{MR14} for understanding CNNs with max-pooling and rectified linear units (ReLUs) through visualization uses deconvolutional neural network~\cite{MGR11} attached to the convolutional network of interest. This approach maps the feature activity in intermediate layers of a previously trained CNN back to the input pixel space using deconvolutional network, which performs successively repeated operations of i) unpooling, ii) rectification, and iii) filtering. Since this method identifies structures within each patch that stimulate a particular feature map, it differs from previous approaches~\cite{girshick2014rcnn} which instead identify patches within a data set that stimulate strong activations at higher layers in the model. The method can also be interpreted as providing an approximation to partial derivatives with respect to pixels in the input image~\cite{DBLP:journals/corr/SimonyanVZ13}. One of the shortcomings of the method is that it enables the visualization of only a single activation in a layer (all other activations are set to zero). There also exist other techniques for inverting a modern large convolutional network with another network, e.g.\ a method based on up-convolutional architecture~\cite{DB16}, where as opposed to the previously described deconvolutional neural network, the up-convolutional network is trained. This method inverts deep image representations and obtains reconstructions of an input image from each layer. 

The fundamental difference between the LRP approach and the deconvolution method lies in how the responses are projected towards the inputs. The latter approach solves the optimization problems to reconstruct the image input while the former one aims to reconstruct the classifier decision (the details are well-explained in~\cite{BachPLOS15}). 

Guided backpropagation~\cite{SprDosBroRied15} extends the deconvolution approach by combining it with a simple technique visualizing the part of the image that most activates a given neuron using a backward pass of the activation of a single neuron after a forward pass through the network. Finally, the recently published method~\cite{DBLP:journals/corr/ZintgrafCW16} based on the prediction difference analysis~\cite{DBLP:journals/tkde/Robnik-SikonjaK08} is a probabilistic approach that extends the idea in~\cite{MR14} of visualizing the probability of the correct class using the occlusion of the parts of the image. The approach highlights the regions of the input image of a CNN which provide evidence for or against a certain class. 

Understanding CNNs can also be done by visualizing output units as distributions in the input space via output unit sampling~\cite{Hinton:2006:FLA:1161603.1161605}. However, computing relevant statistics of the obtained distribution is often difficult. This technique cannot be applied to deep architectures based on auto-encoders as opposed to the subsequent work~\cite{Erhan-vis-techreport-2010,visualization_techreport}, where the authors visualize what is activated by the unit in an arbitrary layer of a CNN in the input space (of images) via an activation maximization procedure that looks for input patterns of a bounded norm that maximize the activation of a given hidden unit using gradient ascent. This method extends previous approaches~\cite{DBLP:journals/neco/BerkesW06}. The gradient-based visualization method~\cite{visualization_techreport} can also be viewed as a generalization of the deconvolutional network reconstruction procedure~\cite{MR14} as shown in subsequent work~\cite{DBLP:journals/corr/SimonyanVZ13}. The requirement of careful initialization limits the method~\cite{MR14}. The approach was applied to Stacked Denoising Auto-Encoders, Deep Belief Networks and later on to CNNs~\cite{DBLP:journals/corr/SimonyanVZ13}. Finally, sensitivity-based methods~\cite{DBLP:journals/corr/SimonyanVZ13,Baehrens:2010:EIC:1756006.1859912,Rasmussen_visualizationof}) aim to understand how the classifier works in different parts of the input domain by computing scores based on partial derivatives at the given sample. 

Some more recent gradient-based visualization techniques for CNN-based models not mentioned before include Grad-CAM~\cite{DBLP:journals/corr/SelvarajuDVCPB16}, which is an extension of the Class Activation Mapping (CAM) method~\cite{cvpr2016_zhou}. The approach heavily relies on the construction of weighted sum of the feature maps, where the weights are global-average-pooled gradients obtained through back-propagation. The approach lacks the ability to show fine-grained importance like pixel-space gradient visualization methods~\cite{SprDosBroRied15,MR14} and thus in practice has to be fused with these techniques to create high-resolution class-discriminative visualizations. 

Finally, complementary bibliography related to analyzing neural networks is briefly discussed in the Supplement (Section~\ref{sec:cb}).

\section{Visualization method}
\label{sec:vism}

As mentioned before, our method combines feature maps from deep convolutional layers that contain mostly relevant information, but are low-resolution, with the feature maps of the shallow layers that have higher resolution but also contain more irrelevant information. This is done by ``back-propagating'' the information about the regions of relevance while simultaneously increasing the resolution. The back-propagation is value-based. We call this approach VisualBackProp to emphasize that we ``back-propagate'' values (images) instead of gradients. We explain the method in details below.

\begin{figure}[!htp]
\vspace{-0.05in}
\centering
a) \includegraphics[width=0.53\columnwidth]{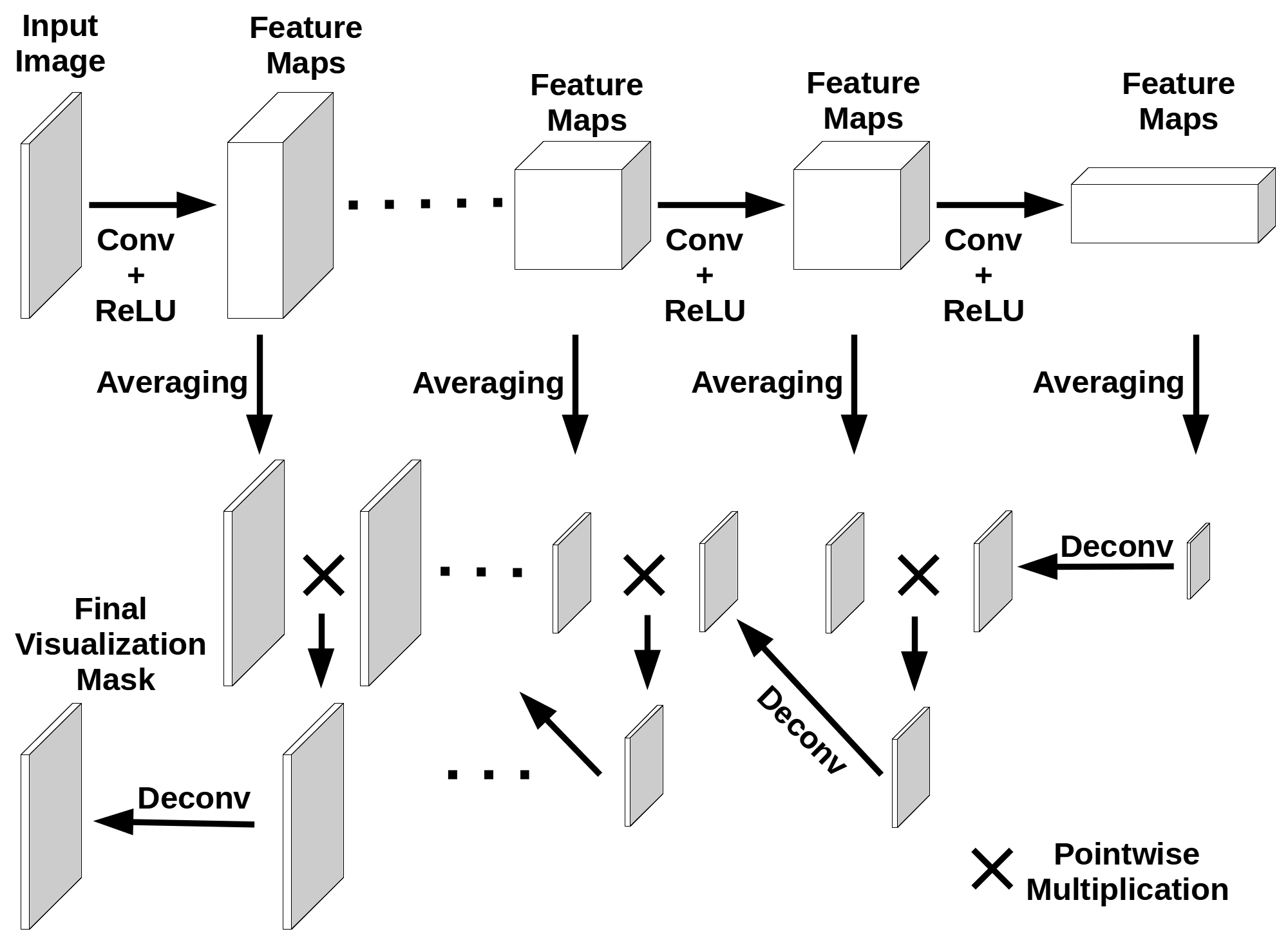} 
\hspace{0.1in}
b) \includegraphics[width=0.19\columnwidth]{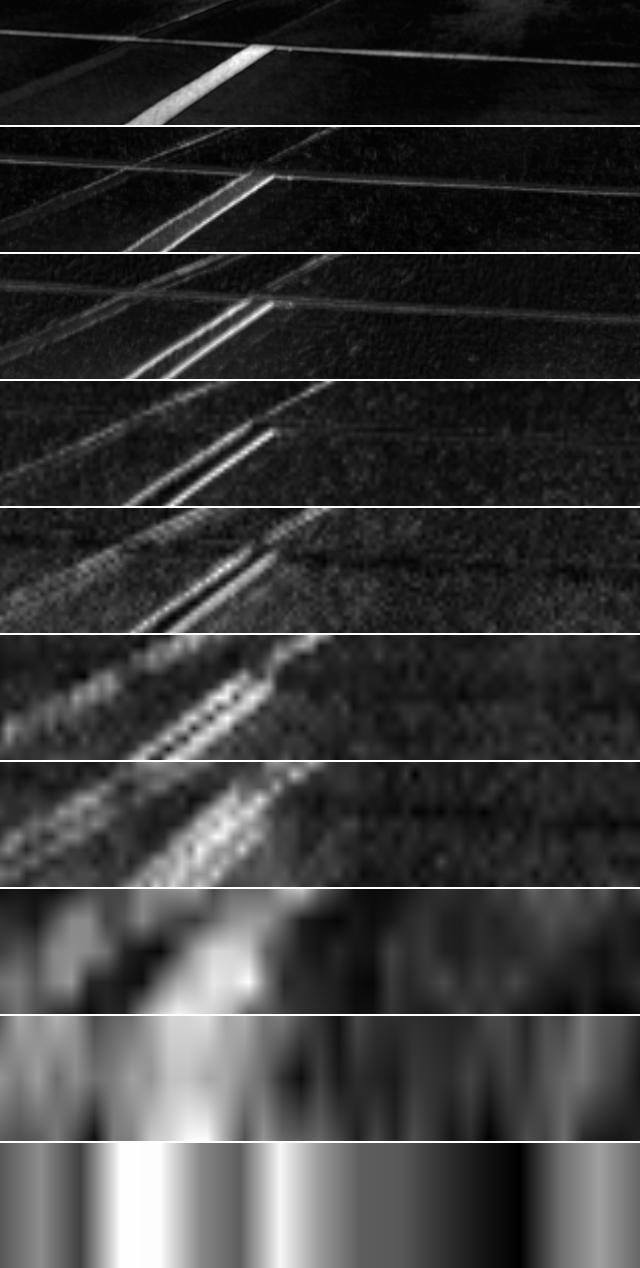}
\includegraphics[width=0.19\columnwidth]{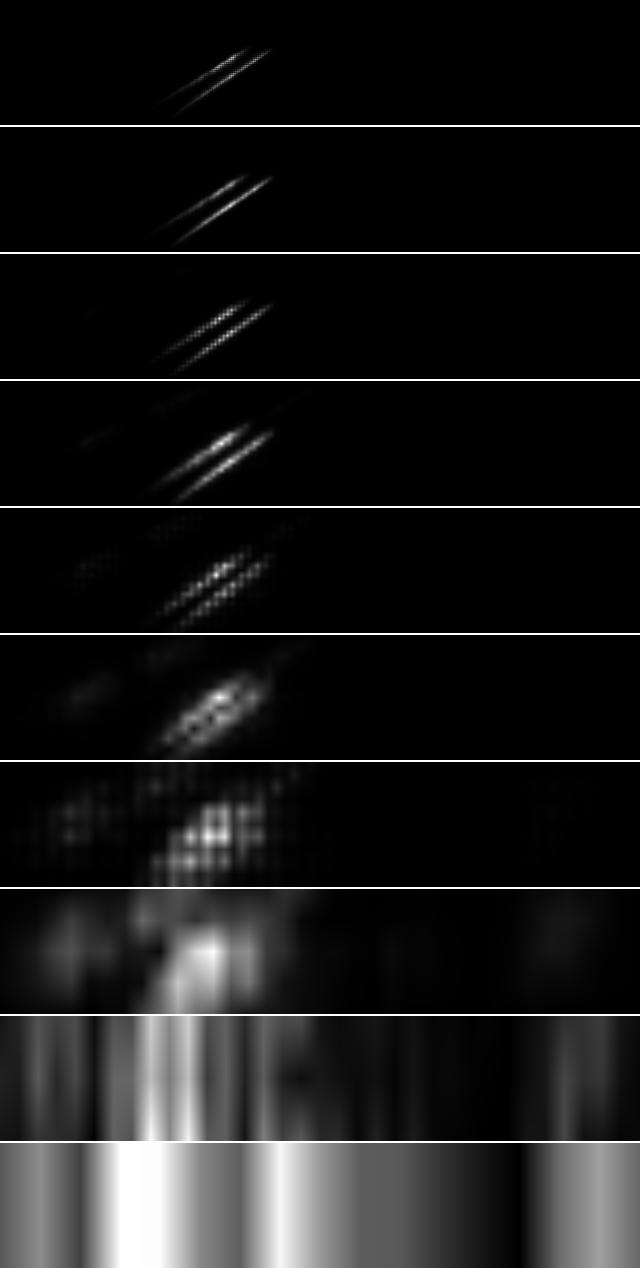}\\
\vspace{-0.05in}
\caption{\textbf{a)} Block diagram of the VisualBackProp method. The feature maps after each ReLU layer are averaged. The averaged feature map of the last convolutional layer is scaled-up via deconvolution and multiplied by the averaged feature map from the previous layer. The resulting intermediate mask is again scaled-up and multiplied. This process is repeated until we reach the input. \textbf{b)} \textit{Left}: Averaged feature maps of all convolutional layers from the input (top) to the output (bottom). \textit{Right}: Corresponding intermediate masks. The top one is the final result.}
\label{vis_diagram}
\vspace{-0.15in}
\end{figure}

The block diagram of the proposed visualization method is shown in Figure~\ref{vis_diagram}a. The method utilizes the forward propagation pass, which is already done to obtain a prediction, i.e. we do not add extra forward passes. The method then uses the feature maps obtained after each ReLU layer (thus these feature maps are already thresholded). In the first step, the feature maps from each layer are averaged, resulting in a single feature map per layer. Next, the averaged feature map of the deepest convolutional layer is scaled up to the size of the feature map of the previous layer. This is done using deconvolution with filter size and stride that are the same as the ones used in the deepest convolutional layer (for deconvolution we always use the same filter size and stride as in the convolutional layer which outputs the feature map that we are scaling up with the deconvolution). In deconvolution, all weights are set to $1$ and biases to $0$. The obtained scaled-up averaged feature map is then point-wise multiplied by the averaged feature map from the previous layer. The resulting image is again scaled via deconvolution and multiplied by the averaged feature map of the previous layer exactly as described above. This process continues all the way to the network's input as shown in Figure~\ref{vis_diagram}a. In the end, we obtain a mask of the size of the input image, which we normalize to the range $[0,1]$.

The process of creating the mask is illustrated in Figure~\ref{vis_diagram}b. On the left side the figure shows the averaged feature maps of all the convolutional layers from the input (top) to the output (bottom). On the right side it shows the corresponding intermediate masks. Thus on the right side we show step by step how the mask is being created when moving from the network's output to the input. Comparing the two top images clearly reveals that many details were removed in order to obtain the final mask. 

\section{Theoretical analysis}
\label{sec:theory}

\begin{figure*}[htp!]
\vspace{-0.1in}
  \center
\includegraphics[width = 0.45\textwidth]{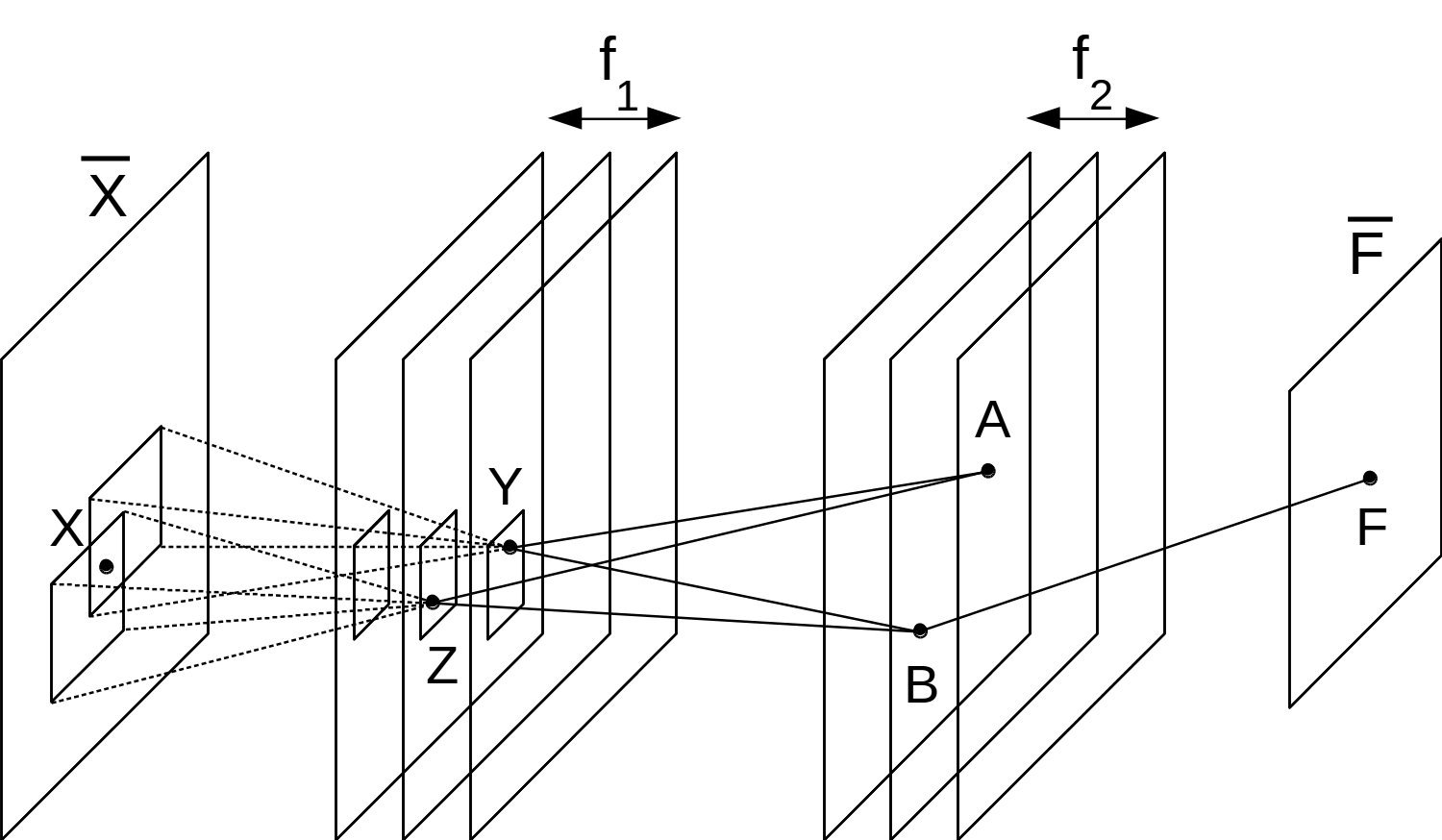}
\hspace{0.1in}\includegraphics[width = 0.45\textwidth]{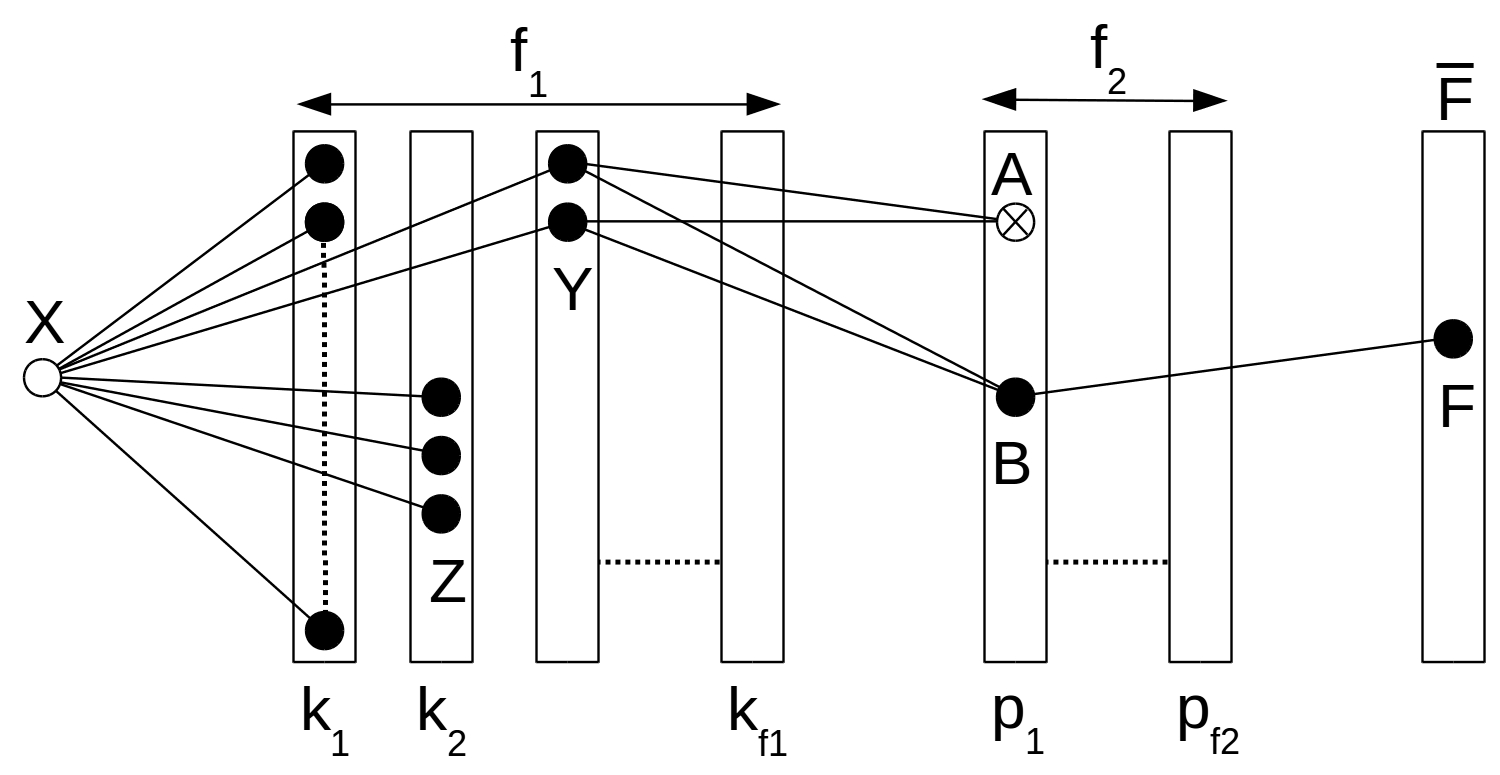}
\vspace{-0.05in}
\caption{\textbf{Left}: General CNN $\mathcal{N}$ with biases. \textbf{Right}: Corresponding multipartite graph. Dead nodes, i.e.\ nodes that do not transmit any flow (corresponding to neurons with zero activation), are crossed.}
\label{fig:fig4}
\vspace{-0.05in}
\end{figure*}

We now present theoretical guarantees for the algorithm (all proofs are deferred to the Supplement). Our theoretical analysis does not rely on computing the sensitivity of any particular cost function with respect to the changes of values of particular input neurons. So we will not focus on computing the gradients. The reason for that is that even if the gradients are large, the actual contribution of the neuron might be small. Instead our proposed method is measuring the actual contribution that takes into account the ``collaborative properties'' of particular neurons. This is measured by the ability of particular neurons to substantially participate in these weighted inputs to neurons in consecutive layers that themselves have higher impact on the form of the ultimate feature maps than others.

Let's consider a convolutional neural network $\mathcal{N}$ with $L$ convolutional layers that we index from $1$ to $L$ and ReLU nonlinear mapping with biases $b(v)$, where $v$ stands for a neuron of the network. We assume that no pooling mechanism is used and the strides are equal to one (the entire analysis can be repeated for arbitrary stride values).
We denote by $f_{l}$ the number of feature maps of the $l^{th}$ layer, by $m_{l} \times r_{l}$ the shape of the kernel applied in the $l^{th}$ convolutional layer. 

For a CNN with learned weights and biases we think about the inference phase as a network flow problem on the sparse multipartite graph with $L+1$ parts $A_{0}, A_{1},...,A_{L}$ ($A_{0}$ is the input image), where different parts $A_{i}$ simply correspond to different convolutional layers, source nodes are the pixels of the input image ($l=0$) and sinks are the nodes of the last convolutional layer. Figure~\ref{fig:fig4} explains the creation of the multipartite graph from the CNN: every vertex of the graph corresponds to a neuron and the number of parts of the graph corresponds to the number of network layers. A vertex v in each part is connected with the vertex in a subsequent part if i) there is an edge between them in the corresponding CNN and ii) this vertex (neuron) v has non-zero activation. Furthermore, we denote as $\gamma(v)$ the incoming flow to node $v$. From now on we will often implicitly transition from the neural network description to the graph theory and vice versa using this mapping. 

Let us now define an important notation we will use throughout the paper:

\begin{itemize}
\vspace{-0.1in}
\item Let $\gamma(v)$ be the value of the total input flow to neuron $v$ (or in other words the weighted input to neuron $v$), $a(v)$ be the value of the activation of $v$, and $b(v)$ be the bias of $v$.
\vspace{-0.05in}
\item Let $e$ be an edge from some neuron $v^{'}$ to $v$. Then $\gamma_e$ will denote the input flow to $v$ along edge $e$, $a_e$ will denote the activation of $v$ (i.e. $a_e \coloneqq a(v)$), and $b_e$ be the bias of $v$ (i.e. $b_e \coloneqq b(v)$).
\vspace{-0.1in}
\end{itemize}

We will use either of the two notations depending on convenience.

Note that the flow is (dis-)amplified when it travels along edges and furthermore some fraction of the flow is lost in certain nodes of the network (due to biases). This is captured in the Lemma~\ref{lem:one}.

\begin{lemma}
\vspace{-0.05in}
The inference phase in the convolutional neural network $\mathcal{N}$ above is equivalent to the network flow model on a multipartite graph with parts $A_{1},...,A_{L}$, where each edge has a weight defining the amplification factor for the flow traveling along this edge. Consider node $v$ in the part $A_{i-1}$. It has $m_{i}r_{i}f_{i}$ neighbors in part $A_{i}$. The flow of value $\min(\gamma(v),b(v))$ is lost in $v$.
\label{lem:one}
\vspace{-0.05in}
\end{lemma}

\begin{figure}[htp!]
\vspace{-0.05in}
  \center
a) \includegraphics[width = 0.38\textwidth]{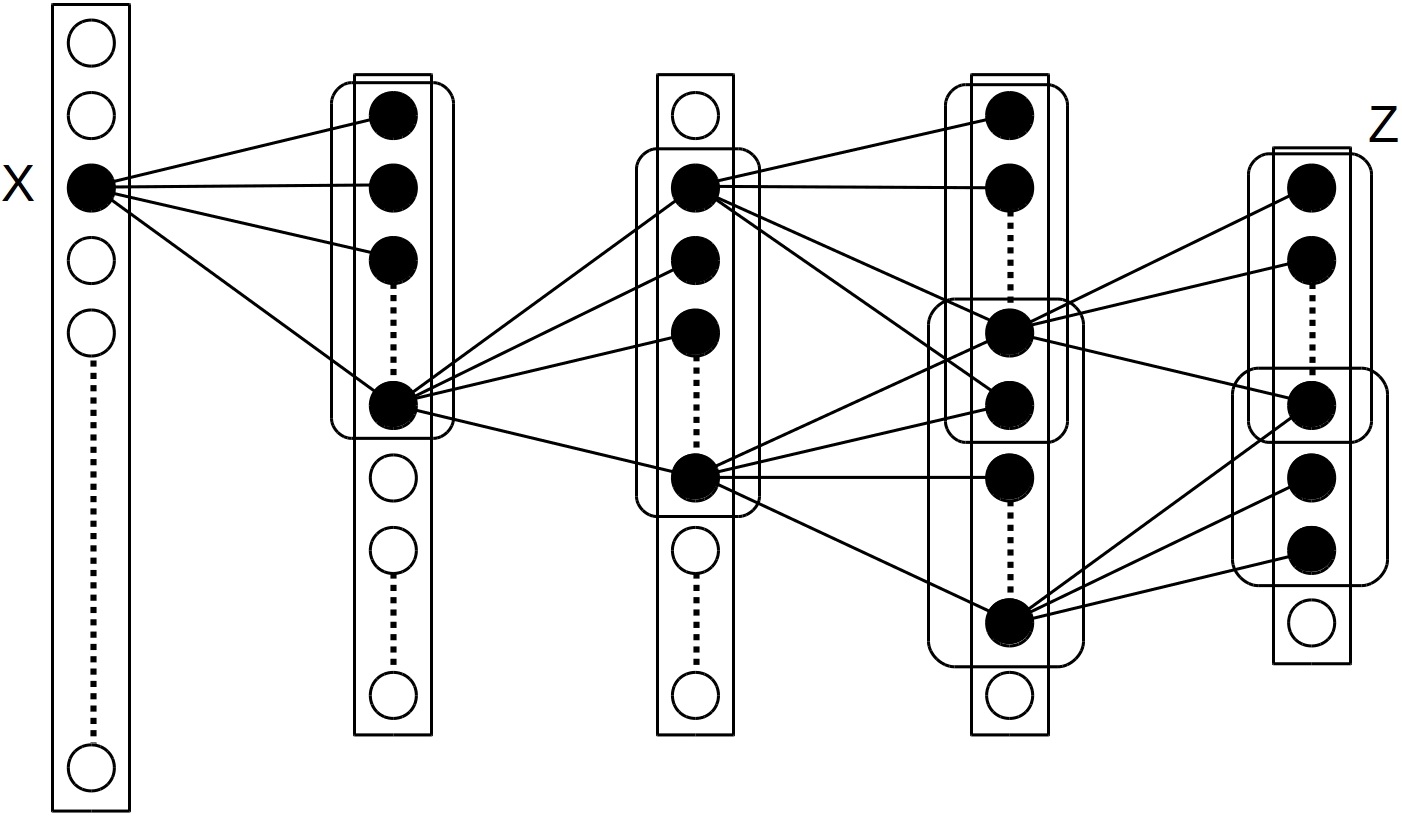}
\includegraphics[width = 0.13\textwidth]{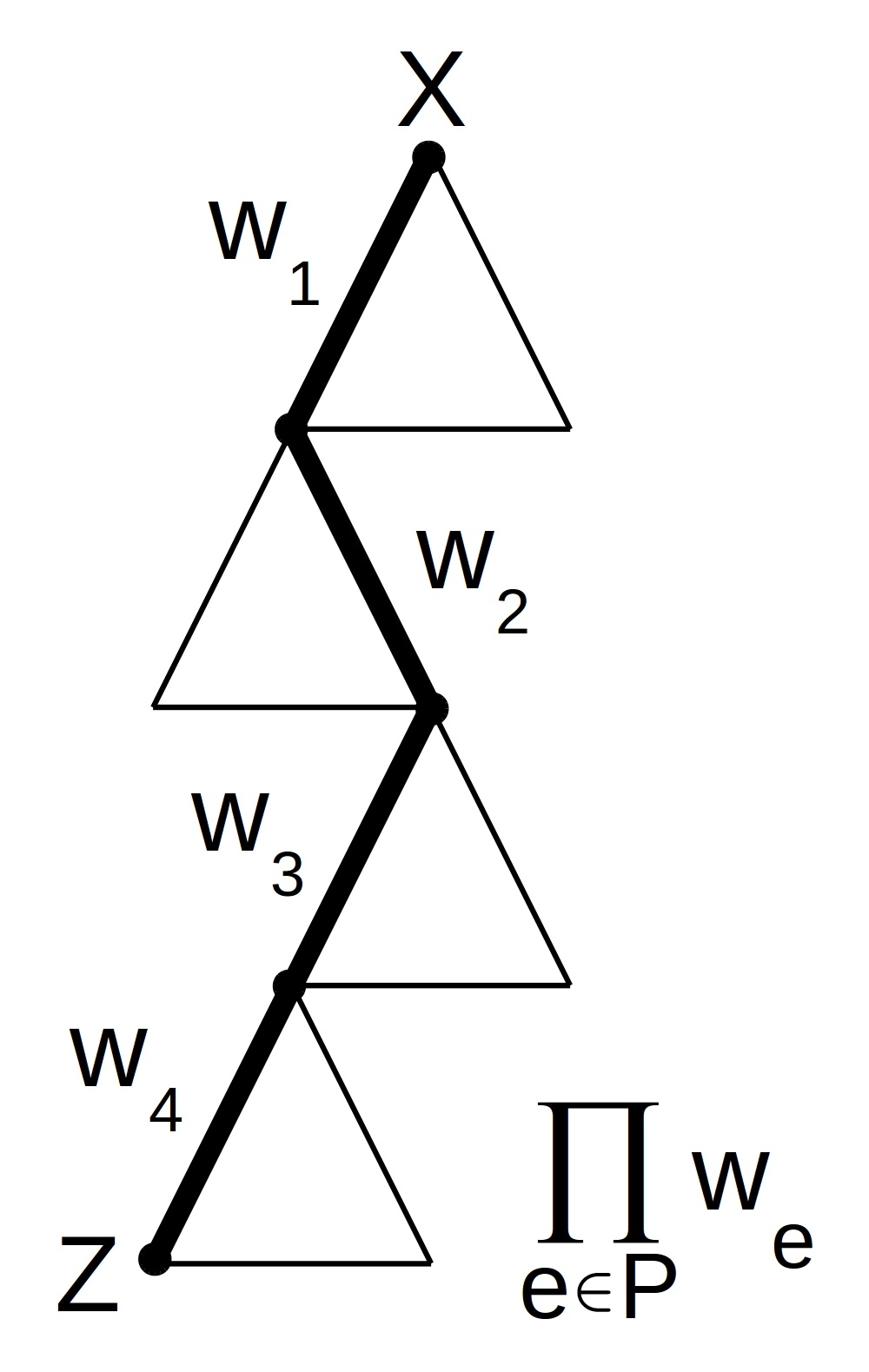}
b) \includegraphics[width = 0.42\textwidth]{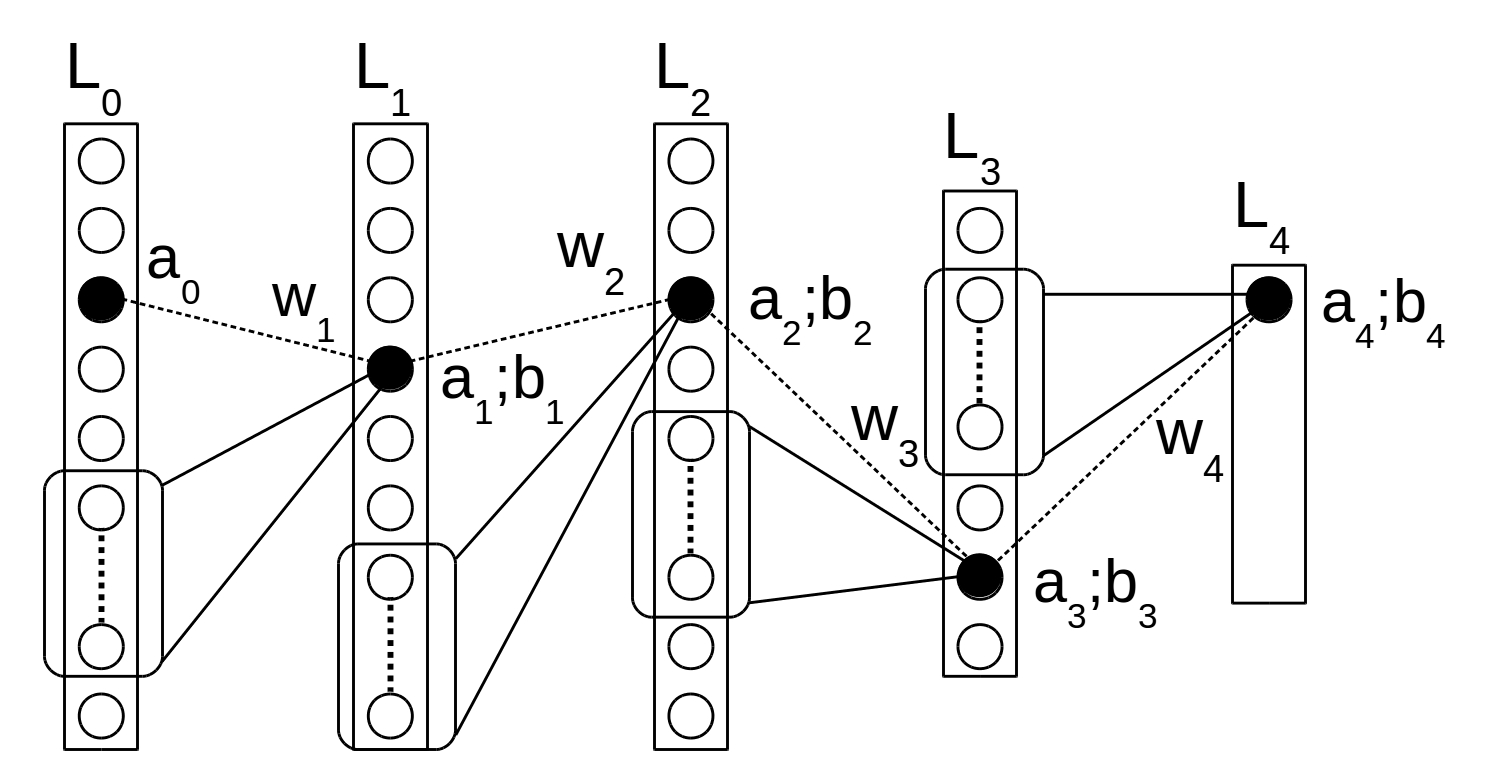}\\
\vspace{-0.05in}
\caption{\textbf{a)} Convolutional architecture with no biases. The contribution of the input pixel $X$ to the last layer of feature maps can be computed by summing the contribution along different paths originating at $X$ and ending in the last layer. The contribution from each path is proportional to the product of the weights of edges belonging to this path. \textbf{b)} The contribution of any given input pixel $X$ along a particular path $P$ in the general CNN $\mathcal{N}$ is equal to the product $\prod_{e \in P} \frac{a_{e}w_{e}}{a_{e}+b_{e}}$, where $a_{e}$s are the activations on the path and $b_{e}$s are the corresponding biases.}
\label{fig:fig3}
\vspace{-0.05in}
\end{figure}

The flow-formulation of the problem is convenient since it will enable us to quantitatively measure the contribution of pixels from the input image to the activations of neurons in the last convolutional layer. Before we propose a definition of that measure, let us consider a simple scenario, where no nonlinear mapping is applied. That neural network model would correspond to the network flow for multipartite graphs described above, but with no loss of the flow in the nodes of the network. In that case different input neurons act independently. For each node $X$ in $A_{0}$ (i.e. $X$ corresponds to a pixel in the input image), the total flow value received by $A_{L}$ is obtained by summing over all paths $P$ from $X$ to $A_{L}$ the expressions of the form $\gamma(X) \prod_{e \in P} w_{e}$, where $w_{e}$ stands for the weight of an edge $e$ and $\gamma(X)$ is the value of pixel $X$ or in other words the value of the input flow to X. This is illustrated in Figure~\ref{fig:fig3}a. This observation motivates the following definition:

\begin{definition}[$\phi$-function for CNNs with no biases]
\vspace{-0.05in}
Consider the neural network architecture, similar to the one described above, but with no biases. Lets call it $\tilde{\mathcal{N}}$. Then the contribution of the input pixel $X$ to the last layer of feature maps is given as:
\begin{equation}
\phi^{\tilde{\mathcal{N}}}(X) \coloneqq \gamma(X)\sum_{P \in \mathcal{P}} \prod_{e \in P} w_{e},
\end{equation}
where $\gamma(X)$ is the value of pixel $X$ and $\mathcal{P}$ is a family of paths from $X$ to $A_{L}$ in the corresponding multipartite graph.
\vspace{-0.05in}
\label{def:one}
\end{definition}

Definition~\ref{def:one} however does not take into consideration network biases. In order to generalize it to arbitrary networks, we construct a transformation of a CNN with biases $\mathcal{N}$ to the one without biases $\tilde{\mathcal{N}}$ as given in Lemma~\ref{lem:two}

\begin{lemma}
\vspace{-0.05in}
The bias-free image $\tilde{\mathcal{N}}$ of the network $\mathcal{N}$ can be obtained from $\mathcal{N}$ by removing the biases and multiplying each weight $w_{e}$ of an edge $e$ by $c_e = \frac{a(v^{'})}{\gamma(v)}$, where an edge $e$ goes from neuron $v^{'}$ to $v$. Note that $\gamma_e = w_ea(v^{'})$.
\label{lem:two}
\vspace{-0.05in}
\end{lemma}

The in-depth discussion of this transformation can be found in the Supplement.

Since we know how to calculate the contribution of each input neuron to the convolutional network in the setting without biases and we know how to translate any general convolutional neural network $\mathcal{N}$ to the equivalent one $\tilde{\mathcal{N}}$ without biases, we are ready to give a definition of the contribution of each input neuron in the general network $\mathcal{N}$.

\begin{definition}[$\phi$-function for general CNNs]
\vspace{-0.1in}
Consider the general neural network architecture described above. Then the contribution of the input pixel $X$ to the last layer of feature maps is given as:
\vspace{-0.1in}
\begin{equation}
\phi^{\mathcal{N}}(X) \coloneqq \phi^{\tilde{\mathcal{N}}}(X),
\end{equation}
\vspace{-0.2in}

\noindent where $\tilde{\mathcal{N}}$ is the bias-free image of network $\mathcal{N}$.
\label{def:two}
\end{definition}

Definition~\ref{def:two} relates the contribution of the neuron $X$ in $\mathcal{N}$ from the parameters of $\tilde{\mathcal{N}}$, rather than explicitly in terms of the parameters of $\mathcal{N}$. This is fixed in the lemma below (Lemma~\ref{lem:three}).

\begin{lemma}
\vspace{-0.05in}
For an input neuron $X$ function $\phi^{\mathcal{N}}(X)$ is defined as:
\vspace{-0.05in}
\begin{equation}
\phi^{\mathcal{N}}(X) = \gamma(X) \sum_{P \in \mathcal{P}} \prod_{e \in P} \frac{a_{e}}{a_{e}+b_{e}} w_{e},
\end{equation}
\vspace{-0.15in}

where $\gamma(X)$ is the value of pixel $X$ and $\mathcal{P}$ is a family of paths from $X$ to $A_{L}$ in the corresponding multipartite graph. 
\label{lem:three}
\end{lemma}

Figure~\ref{fig:fig3}b illustrates the lemma. Note that in the network with no biases one can set $b_{e}=0$ and obtain the results introduced before the formula involving sums of weights' products.

We prove the following results (borrowing the notation introduced before) by connecting our previous theoretical analysis with the algorithm.

\begin{theorem}
\vspace{-0.05in}
For a fixed CNN $\mathcal{N}$ considered in this paper there exists a universal constant $c>0$ such that the values of the input neurons computed by \textit{VisualBackProp}
are of the form:
\vspace{-0.05in}
\begin{equation}
\phi^{\mathcal{N}}_{VBP}(X) = c \cdot \gamma(X) \sum_{P \in \mathcal{P}} \prod_{e \in P} a_{e}.
\end{equation}
\label{thm:four}
\vspace{-0.2in}
\end{theorem}

The statement above shows that the values computed for pixels by the VisualBackProp algorithm are related to the flow contribution from that pixels in the corresponding graphical model and thus, according to our analysis, measure their importance. The formula on $\phi^{\mathcal{N}}_{VBP}(X)$ is similar to the one on $\phi^{\mathcal{N}}(X)$, but gives rise to a much more efficient algorithm and leads to tractable theoretical analysis. Note that the latter one can be obtained from the former one by multiplying each term of the inner products by $\frac{w_{e}}{a_{e}+b_{e}}$ and then rescaling by a multiplicative factor of $\frac{1}{c}$.
Rescaling does not have any impact on quality since it is conducted in exactly the same way for all the input neurons. Finally, the following observation holds.

\begin{remark}
\vspace{-0.05in}
Note that for small kernels the number of paths considered in the formula on $\phi^{\mathcal{N}}(X)$ is small (since the degrees of the corresponding multipartite graph are small) thus in practice the difference between formula on $\phi^{\mathcal{N}}_{VBP}(X)$ and the formula on $\phi^{\mathcal{N}}(X)$ coming from the re-weighting factor $\frac{w_{e}}{a_{e}+b_{e}}$ is also small. Therefore for small kernels the VisualBackProp algorithm computes good approximations of input neurons' contributions to the activations in the last layer.
\end{remark}

In the next section we show empirically that $\phi^{\mathcal{N}}_{VBP}(X)$ works very well as a measure of contribution\footnote{$\phi^{\mathcal{N}}(X)$ thus might be a near-monotonic function of $\phi^{\mathcal{N}}_{VBP}(X)$. We leave studying this to future works.}. 

\section{Experiments}
\label{sec:exper}

In the main body of the paper we first demonstrate the performance of VisualBackProp on the task of end-to-end autonomous driving, which requires real-time operation. The codes of VisualBackProp are already publicly released. The experiments were performed on the Udacity self-driving car data set (Udacity Self Driving Car Dataset 3-1: El Camino). We qualitatively compare our method with LRP implementation as given in Equation 6 from~\cite{DBLP:journals/corr/SamekBMBM15} (similarly to the authors, we use $\epsilon = 100$) and we also compare their running times\footnote{Note that it is hard to quantitatively measure the performance of the visualization methods and none such standardized measures exist.}. We then show experimental results on the task of the classification of traffic signs on the German Traffic Sign Detection Benchmark data set (\url{http://benchmark.ini.rub.de/?section=gtsdb&subsection=dataset}) and ImageNet data set (\url{http://image-net.org/challenges/LSVRC/2016/}). Supplement contains additional experimental results on all three data sets. 

We train two networks, that we call \textit{NetSVF} and \textit{NetHVF}, which vary in the input size. In particular, \textit{NetHVF} input image has approximately two times higher vertical field of view, but then is scaled down by that factor. The details of both architectures are described in Table~\ref{tab:nnarch} in the Supplement. The networks are trained with stochastic gradient descent (SGD) and the mean squared error (MSE) cost function for $32$ epochs.  

\begin{figure}[t!]
\centering
\includegraphics[width=0.245\columnwidth]{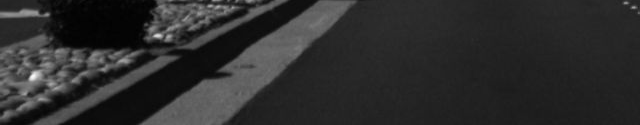}
\includegraphics[width=0.245\columnwidth]{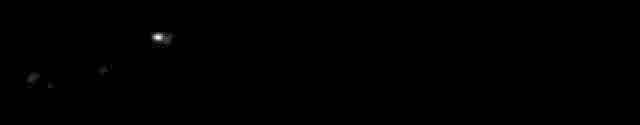} 
\includegraphics[width=0.245\columnwidth]{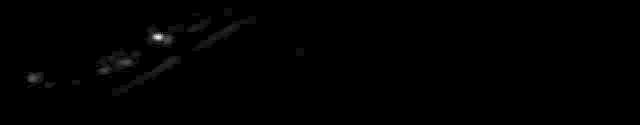} 
\includegraphics[width=0.245\columnwidth]{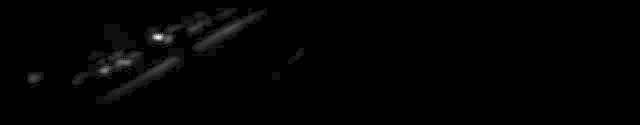} \\
\vspace{0.02in}  
\includegraphics[width=0.245\columnwidth]{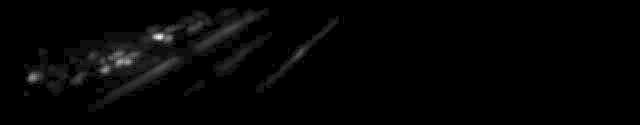}  
\includegraphics[width=0.245\columnwidth]{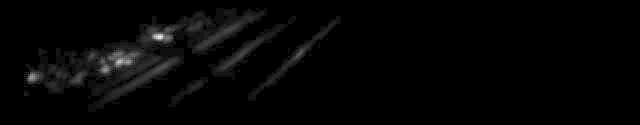}  
\includegraphics[width=0.245\columnwidth]{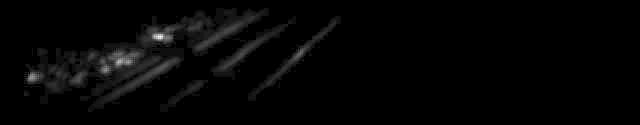}
\includegraphics[width=0.245\columnwidth]{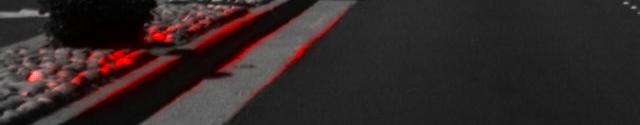} \\
\vspace{-0.1in}
\caption{Visualizing the training of the network \textit{NetSVF} with the VisualBackProp method: (starting from the second image in the first row) the mask after 0.1, 0.5, 1, 4, 16, and 32 epochs (similar figure for \textit{NetHVF} is shown in the Supplement). Over time the network learns to detect more relevant cues for the task from the training data. The first image of the first row is an input image and the last image of the second row is the input image with the mask overlaid in red.} 
\label{exp_f6a}
\vspace{-0.2in}
\end{figure}

The Udacity self-driving car data set that we use contains images from three front-facing cameras (left, center, and right) and measurements such as speed and steering wheel angle, recorded from the vehicle driving on the road. The measurements and images are recorded with different sampling rates and are independent, thus they require synchronization before they can be used for training neural networks. For the purpose of this paper, we pre-process the data with the following operations: i) synchronizing images with measurements from the vehicle, ii) selecting only center camera images, iii) selecting images where the car speed is above $5$~m/s, iv) converting images to gray scale, and v) cropping and scaling the lower part of the images to a $640\times 135$ size for network \textit{NetSVF} and $351\times 135$ size for network \textit{NetHVF}. As a result, we obtain a data set with $210$~K images with corresponding steering wheel angles (speed is not used), where the first $190$~K examples are used for training and the remaining $20$~K examples are used for testing. We train the CNN to predict steering wheel angles based on the input images.

\begin{figure}[h!]
\vspace{-0.05in}
\centering
\includegraphics[width=0.245\columnwidth]{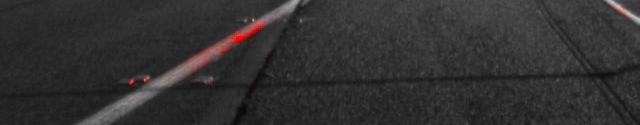}
\includegraphics[width=0.245\columnwidth]{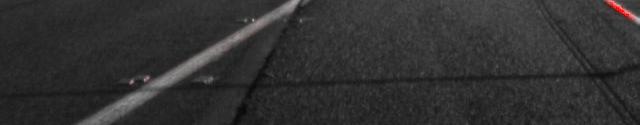} 
\includegraphics[width=0.245\columnwidth]{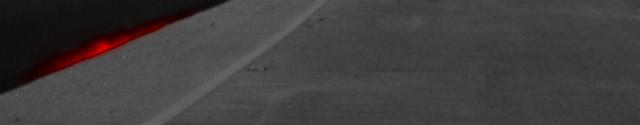}
\includegraphics[width=0.245\columnwidth]{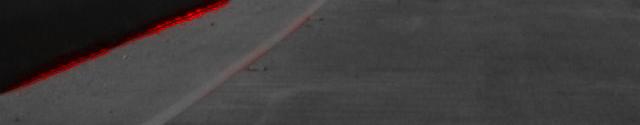} \\
\vspace{0.01in}
\includegraphics[width=0.245\columnwidth]{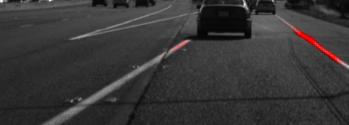}
\includegraphics[width=0.245\columnwidth]{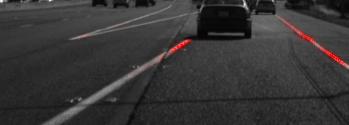} 
\includegraphics[width=0.245\columnwidth]{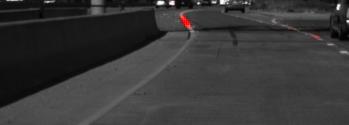}
\includegraphics[width=0.245\columnwidth]{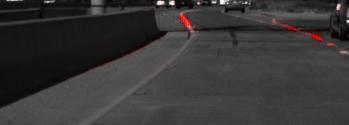} \\
\vspace{-0.05in}
\begin{flushleft}
\hspace{0.2in} VisualBackProp \hspace{0.7in} LRP \hspace{0.8in} VisualBackProp \hspace{0.7in} LRP
\end{flushleft}
\vspace{-0.15in}
\caption{Top: \textit{NetSVF}, bottom: \textit{NetHVF}. Input test images with the corresponding masks overlaid in red. The errors are (from the left): $0.18$ and $-2.41$ degrees of SWA for NetSVF and $0.06$ and $-0.82$ for NetHVF degrees of SWA.} 
\label{exp_f1a}
\vspace{-0.05in}
\end{figure}

\begin{figure}[!htp]
\vspace{-0.12in}
\centering
\includegraphics[width=0.245\columnwidth]{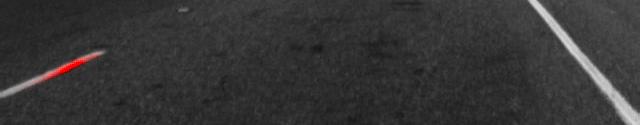}
\includegraphics[width=0.245\columnwidth]{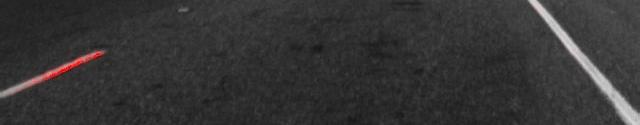} 
\includegraphics[width=0.245\columnwidth]{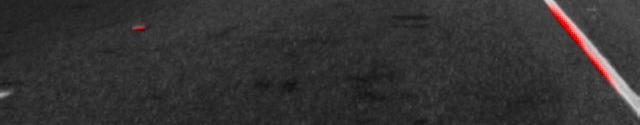}
\includegraphics[width=0.245\columnwidth]{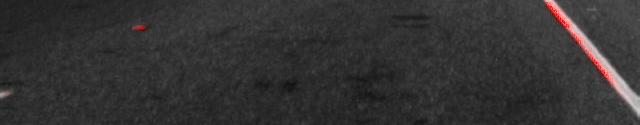}\\
\vspace{-0.05in}
\begin{flushleft}
\hspace{0.2in} VisualBackProp \hspace{0.7in} LRP \hspace{0.8in} VisualBackProp \hspace{0.7in} LRP
\end{flushleft}
\vspace{-0.15in}
\caption{Network \textit{NetSVF}. Two consecutive frames with the corresponding masks overlaid in red. The errors are (from left): $-2.65$ and $3.21$ degrees of SWA.} 
\label{exp_f3a}
\vspace{-0.05in}
\end{figure}

\begin{figure}[htp!]
\vspace{-0.12in}
\centering
\includegraphics[width=0.245\columnwidth]{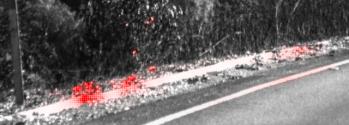}
\includegraphics[width=0.245\columnwidth]{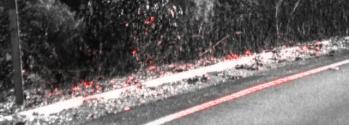} 
\includegraphics[width=0.245\columnwidth]{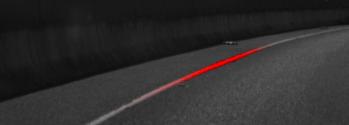}
\includegraphics[width=0.245\columnwidth]{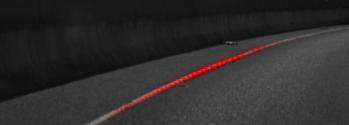}\\
\vspace{-0.05in}
\begin{flushleft}
\hspace{0.2in} VisualBackProp \hspace{0.7in} LRP \hspace{0.8in} VisualBackProp \hspace{0.7in} LRP
\end{flushleft}
\vspace{-0.15in}
\caption{Network \textit{NetHVF}. Input test image with the mask overlaid in red. The errors are (from left): $-20.74$ and $-4.09$ degrees of SWA.} 
\label{exp_f4b}
\vspace{-0.05in}
\end{figure}

\begin{figure}[!htp]
\vspace{-0.05in}
\centering
\includegraphics[width=0.245\columnwidth]{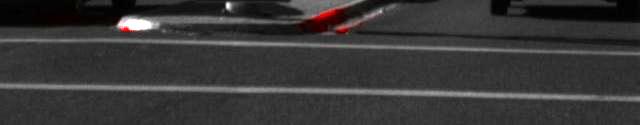}
\includegraphics[width=0.245\columnwidth]{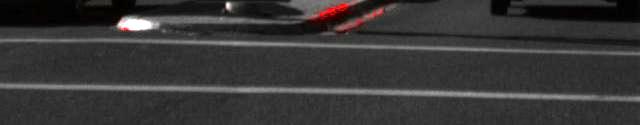} 
\includegraphics[width=0.245\columnwidth]{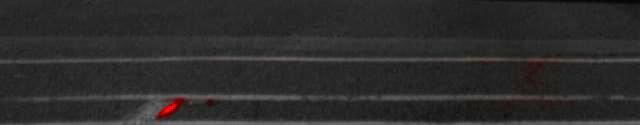}
\includegraphics[width=0.245\columnwidth]{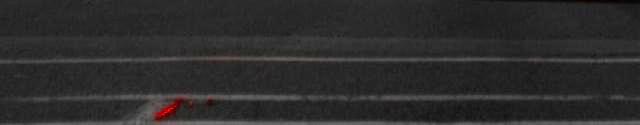} \\
\vspace{-0.05in}
\begin{flushleft}
\hspace{0.2in} VisualBackProp \hspace{0.7in} LRP \hspace{0.8in} VisualBackProp \hspace{0.7in} LRP
\end{flushleft}
\vspace{-0.15in}
\caption{Network \textit{NetSVF}. Input test image with the mask overlaid in red. The errors are (from the top): $-3.28$ and $-0.84$ degrees of SWA.} 
\label{exp_f5a}
\vspace{-0.05in}
\end{figure}

We first apply VisualBackProp method during training to illustrate the development of visual cues that the network focuses on. The obtained masks for an exemplary image are shown in Figure~\ref{exp_f6a}. The figure captures how the CNN gradually learns to recognize visual cues relevant for steering the car (lane markings) as the training proceeds.

We next evaluate VisualBackProp and LRP on the test data. We show the obtained masks on various exemplary test input images in Figures~\ref{exp_f1a}--\ref{exp_f5a} and Figures~\ref{exp_f3b}--\ref{exp_f2b} (Supplement), where on each figure the left column corresponds to our method and the right column corresponds to LRP. For each image we also report the test error defined as a difference between the actual and predicted steering wheel angle (SWA) in degrees. Figures~\ref{exp_f1a} illustrates that the CNN learned to recognize lane markings, the most relevant visual cues for steering a car. It also shows that the field of view affects the visualization results significantly. Figures~\ref{exp_f2a} and~\ref{exp_f2b} capture how the CNN responds to shadows on the image. One can see that the network still detects lane markings but only between the shadows, where they are visible. Each of the Figures~\ref{exp_f3a} and~\ref{exp_f3b} shows two consecutive frames. On the second frame in Figure~\ref{exp_f3a}, the lane marking on the left side of the road disappears, which causes the CNN to change the visual cue it focuses on from the lane marking on the left to the one on the right. Figures~\ref{exp_f4a} and~\ref{exp_f4b} correspond to the sharp turns. The images in the top row of Figure~\ref{exp_f4b} demonstrate the correlation between the high prediction error of the network and the low-quality visual cue it focuses on. Finally, in Figure~\ref{exp_f5a} we demonstrate that the CNN has learned to ignore horizontal lane markings as they are not relevant for steering a car, even though it was trained only with the images and the steering wheel angles as the training signal. The network was therefore able to learn which visual cues are relevant for the task of steering the car from the steering wheel angle alone. Figure~\ref{exp_f5b} similarly shows that the CNN learned to ignore the horizontal lines, however, as the visualization shows, it does not identify lane markings as the relevant visual cues but other cars instead. 

\begin{figure}[htp!]
\vspace{-0.1in}
\centering
\includegraphics[width=0.24\columnwidth]{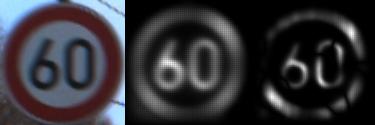}
\includegraphics[width=0.24\columnwidth]{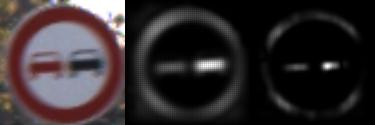}
\includegraphics[width=0.24\columnwidth]{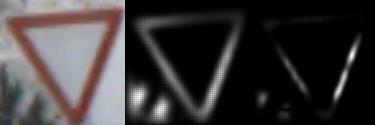}
\includegraphics[width=0.24\columnwidth]{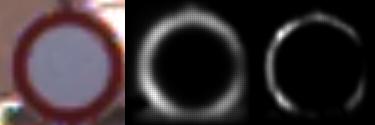}\\
\vspace{0.02in}
\includegraphics[width=0.24\columnwidth]{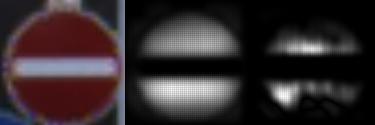}
\includegraphics[width=0.24\columnwidth]{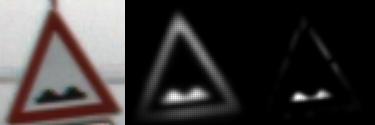}
\includegraphics[width=0.24\columnwidth]{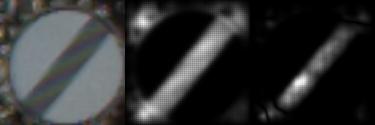}
\includegraphics[width=0.24\columnwidth]{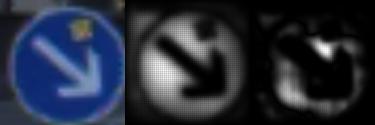}\\
\vspace{-0.05in}
\caption{German Traffic Sign Detection Benchmark data set. Sets of input images with visualization masks arranged in four columns. Each set consist of an input image (left), a visualization mask generated by VisualBackProp (center), and a visualization mask generated by LRP (right).}
\label{signs1}
\vspace{-0.05in}
\end{figure}

\begin{figure}[!htp]
\vspace{-0.1in}
\centering
\includegraphics[width=0.245\columnwidth]{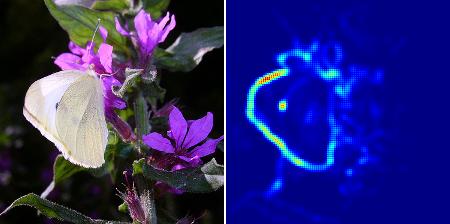} 
\includegraphics[width=0.245\columnwidth]{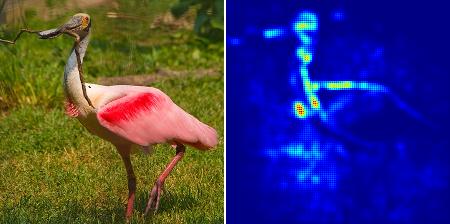}
\includegraphics[width=0.245\columnwidth]{img13.jpeg} 
\includegraphics[width=0.245\columnwidth]{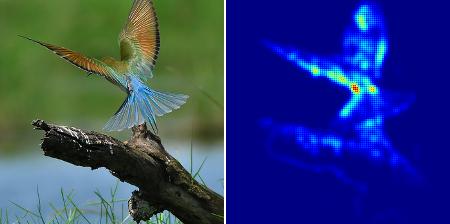}\\
\vspace{0.02in}
\includegraphics[width=0.245\columnwidth]{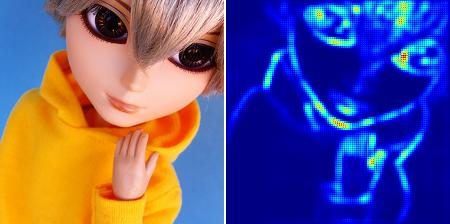} 
\includegraphics[width=0.245\columnwidth]{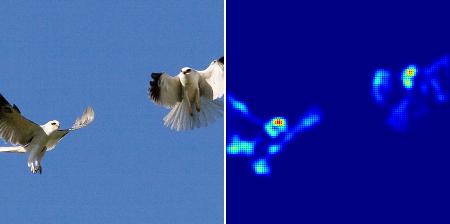}
\includegraphics[width=0.245\columnwidth]{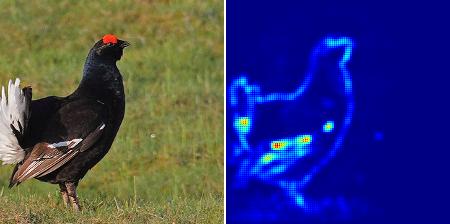} 
\includegraphics[width=0.245\columnwidth]{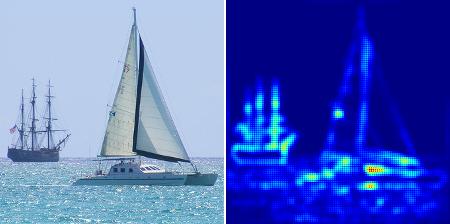}\\
\vspace{-0.05in}
\caption{ImageNet. Sets of input images with visualization masks arranged in four columns. Each set consist of an input image (left), and a visualization mask generated by VisualBackProp (right).}
\label{resnet2}
\vspace{-0.05in}
\end{figure}

We implemented VisualBackProp and LRP in Torch7 to compare the computational time. Both methods used the \textsf{cunn} library to utilize GPU for calculations and have similar levels of optimization. All experiments were performed on GeForce GTX 970M. The average time of computing a mask for VisualBackProp was equal to $2.0\:ms$, whereas in case of the LRP method it was $24.6\:ms$. The VisualBackProp is therefore on average $\bm{12}$ \textbf{times faster} than LRP. At the same time, as demonstrated in Figures~\ref{exp_f1a}--\ref{exp_f5a} and Figures~\ref{exp_f3b}--\ref{exp_f2b} (Supplement), VisualBackProp generates visualization masks that are very similar to those obtained by LRP. 

We next (Figure~\ref{signs1}) demonstrate the performance of VisualBackProp and LRP on German Traffic Sign Detection Benchmark data set. The network is described in Table~\ref{tab:nnarch2} in the Supplement.

Finally, we show the performance of VisualBackProp on ImageNet data set. The network here is a ResNet-$200$~\cite{He2016IdentityMI}.

\vspace{-0.05in}
\section{Conclusions}
\label{sec:concl}
In this paper we propose a new method for visualizing the regions of the input image that have the highest influence on the output of a CNN. The presented approach is computationally efficient which makes it a feasible method for real-time applications as well as for the analysis of large data sets. We provide theoretical justification for the proposed method and empirically show on the task of autonomous driving that it is a valuable diagnostic tool for CNNs.

\small{
\bibliographystyle{unsrt}
\bibliography{VisualBackProp}
}

\normalsize

\clearpage
\newpage

\vbox{\hsize\textwidth
\linewidth\hsize \vskip 0.1in \toptitlebar \centering
\LARGE\bf VisualBackProp: efficient visualization of CNNs\\(Supplementary Material)
\par}  
\bottomtitlebar

\section{Complementary bibliography}
\label{sec:cb}

Other approaches for analyzing neural networks not mentioned in the main body of the paper include quantifying variable importance in neural networks [29, 30], extracting the rules learned by the decision tree model that is fitted to the function learned by the neural network [31], applying kernel analysis to understand the layer-wise evolution of the representation in a deep network [32], analyzing the visual information in deep image representations by looking at the inverse representations [33], applying contribution propagation technique to provide per-instance
explanations of predictions [34] (the method relies on the technique of [35], or visualizing particular neurons or neuron layers [2, 36]. Finally, there also exist more generic tools for explaining individual classification decisions of any classification method for single data instances, like for example~\cite{Baehrens:2010:EIC:1756006.1859912}.

Complementary references:

\small{
\begin{itemize}
\item[] [29] M. Gevrey, I. Dimopoulos, and S. Lek. Review and comparison of methods to study the contribution of variables in artificial neural network models. Ecological Modelling, 160(3):249 – 264, 2003.
\item[] [30] J. D. Olden, M. K. Joy, and R. G. Death. An accurate comparison of methods for quantifying variable importance in artificial neural networks using simulated data. Ecological Modelling, 178(3–4):389 – 397, 2004.
\item[] [31] R. Setiono and H. Liu. Understanding neural networks via rule extraction. In IJCAI, 1995.
\item[] [32] G. Montavon, M. L. Braun, and K.-R. Müller. Kernel analysis of deep networks. J. Mach. Learn. Res., 12:2563–2581, 2011.
\item[] [33] A. Mahendran and A. Vedaldi. Understanding deep image representations by inverting them. In CVPR, 2015.
\item[] [34] W. Landecker, M.l D. Thomure, L. M. A. Bettencourt, M. Mitchell, G. T. Kenyon, and S. P. Brumby. Interpreting individual classifications of hierarchical networks. In CIDM, 2013.
\item[] [35] B. Poulin, R. Eisner, D. Szafron, P. Lu, R. Greiner, D. S. Wishart, A. Fyshe, B. Pearcy, C. Macdonell, and J. Anvik. Visual explanation of evidence with additive classifiers. In IAAI, 2006.
\item[] [36] J. Yosinski, J. Clune, A. M. Nguyen, T. Fuchs, and H. Lipson. Understanding neural networks through deep visualization. CoRR, abs/1506.06579, 2015.
\end{itemize}
}
\normalsize


\section{Proof of Lemma~\ref{lem:one}}
\begin{proof}

\begin{figure}[htp!]
\vspace{-0.15in}
  \center
\includegraphics[width = 0.7\textwidth]{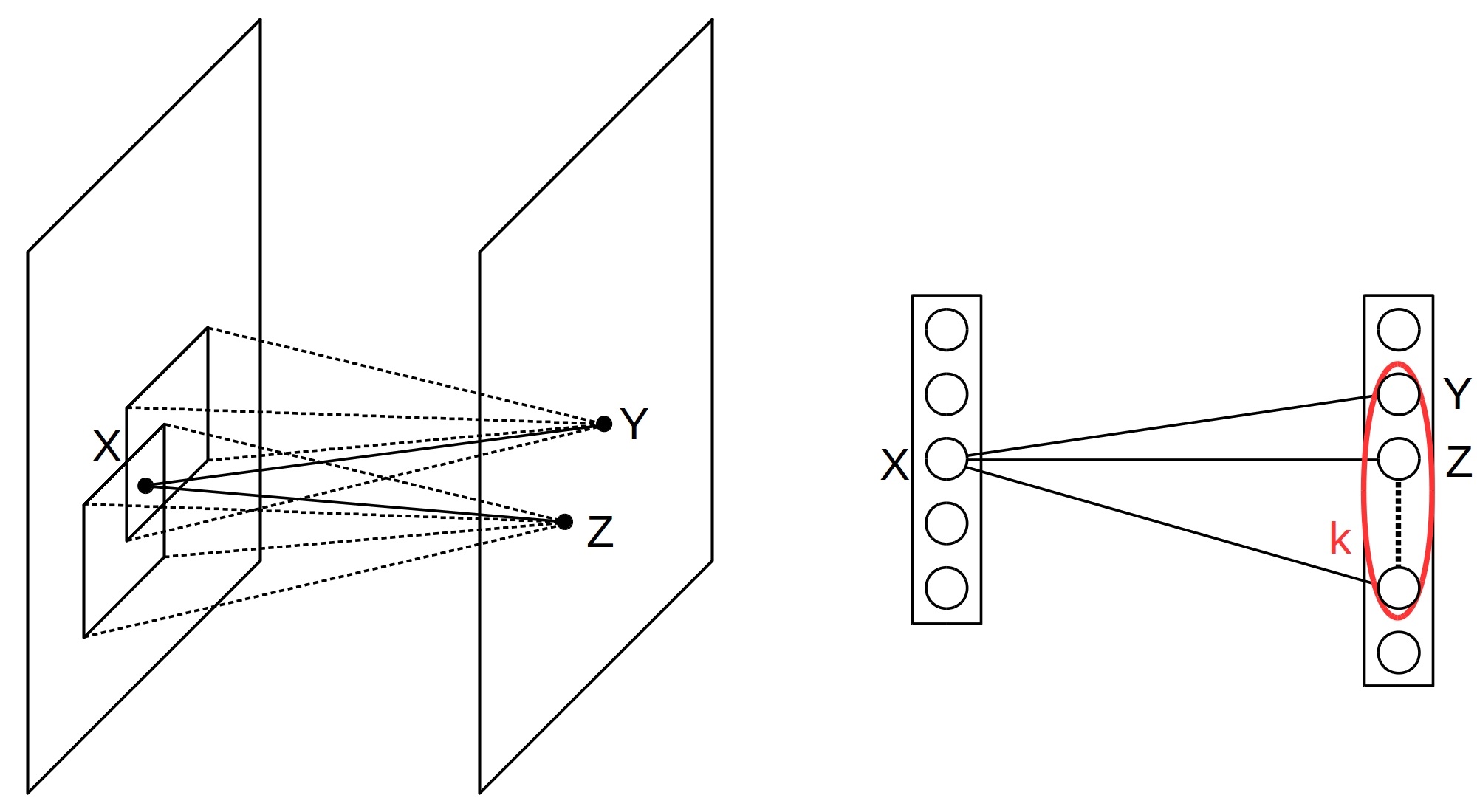}\\
\caption{Input pixel $X$ is covered by patches corresponding to selected neurons in the consecutive layer. In the corresponding multipartite graph node $X$ is adjacent to selected vertices in the next part and transports the flow to them. The same is true if instead of an input pixel we consider any neuron in the network.}
\label{fig:fig1}
\vspace{-0.1in}
\end{figure}

As we have explained in the main body of the paper, we will identify the source-nodes for the flow problem with input pixels and the sink-nodes with the neurons of the last convolutional layer. Parts $A_{1}$,...,$A_{L}$ are the sets of nodes in the corresponding convolutional layers (observe that indeed there are no edges between nodes in $A_{i}$s since there are no edges between neurons in feature maps associated with the fixed convolutional layer). 
Part $A_{0}$ represents the input image. Each source-node $v \in A_{i-1}$ sends $a(v)$ units of flow along edges connecting it with nodes in part $A_{i}$. This is exactly the amount of the input flow $\gamma(v)$ to the node $v$ with flow of value $\min(\gamma(v),b(v))$ deducted (due to the ReLU nonlinear mapping).
The activations of neurons in the last convolutional layer can be represented in that model as the output flows of the nodes from $A_{L}$. 
Note that for a fixed neuron $v$ in the $(i-1)^{th}$ convolutional layer and fixed feature map in the consecutive convolutional layer (here the $0^{th}$ layer is the input image) the kernel of size $m_{i} \times r_{i}$ can be patched in $m_{i}r_{i}$ different ways (see: Figure~\ref{fig:fig1}) to cover neuron $v$ (this is true for all but the ``borderline neurons'' which we can neglect without loss of generality; if one applies padding mechanism the ``borderline neurons'' do not exist at all).
Thus the total number of connections of $v$ to the neurons in the next layer is $m_{i}r_{i}f_{i}$. A similar analysis can be conducted if $v$ is a pixel from the input image.
That completes the proof. 
\end{proof}

\section{Proof of Lemma~\ref{lem:two}}
\begin{proof}
Assume that $e$ is an edge from node $v^{'}$ to $v$. Note that the flow of value $a_{e}\frac{\gamma_e}{\gamma(v)}$ is transmitted by $v$.
Furthermore $\gamma_e = \tilde{\gamma}_e w_{e}$, where $\tilde{\gamma}_e$ is the original value of flow before amplification obtained by passing that flow through an edge $e$.
Thus the flow $\tilde{\gamma}_e$ output by $v^{'}$ is multiplied by $w_{e}\frac{a(v^{'})}{\gamma(v)}$ and output by $v$. Thus one can set: $c(e) = \frac{a(v^{'})}{\gamma(v)}$. That completes the proof.
\end{proof}

\section{Discussion of transformation from Lemma~\ref{lem:two}}

\begin{remark}
The network after transformation is equivalent to the original one, i.e.\ it produces the same activations for all its neurons as the original one.
\end{remark}

\begin{remark}[Intuition behind the transformation:] Let $e$ be an edge from neuron $v^{'}$ to $v$. The transformation replaces the weight of an edge $e$, or the input flow $\gamma_{e}$ along $e$, by $a(v) \frac{\gamma_{e}}{\gamma(v)}$, where $\gamma(v) = \sum_{\bar{e}} \gamma_{\bar{e}}$ is the some of the input flows to $v$ along all the edges going to $v$ and $a(v)$ is an activation of $v$ in the original network. This is done to make the weight of $e$ equal to the relative contribution of $v^{'}$ to the activation of $v$. The relative contribution needs to be on the one hand proportional to the activation of $v$, i.e. $a(v)$, and on the other hand proportional to the input flow along $e$. Finally, the sum of all relative contributions of all the neurons connected with neuron $v$ needs to be equal to the activation $a(v)$. 
\end{remark}

\begin{remark}
The property that the contribution of the given input flow to $v$ to the activation of $v$ is proportional to the value of the flow is a desirable property since the applied nonlinear mapping is a ReLU function. It also captures the observation that if the flow is large, but the activation is small, the contribution might be also small since even though the flow contributes substantially, it contributes to an irrelevant feature.
\end{remark}

\section{Proof of Lemma~\ref{lem:three}}
\begin{proof}
Straightforward from the formula on $\phi^{\tilde{\mathcal{N}}}(X)$ and derived formula on transforming weights of $\mathcal{N}$ to obtain $\tilde{\mathcal{N}}$.
\end{proof}

\newpage
\section{Proof of Theorem~\ref{thm:four}}
\begin{wrapfigure}{l}{0.55\textwidth} 
\vspace{-0.2in}
  \center
\includegraphics[width = 0.25\textwidth]{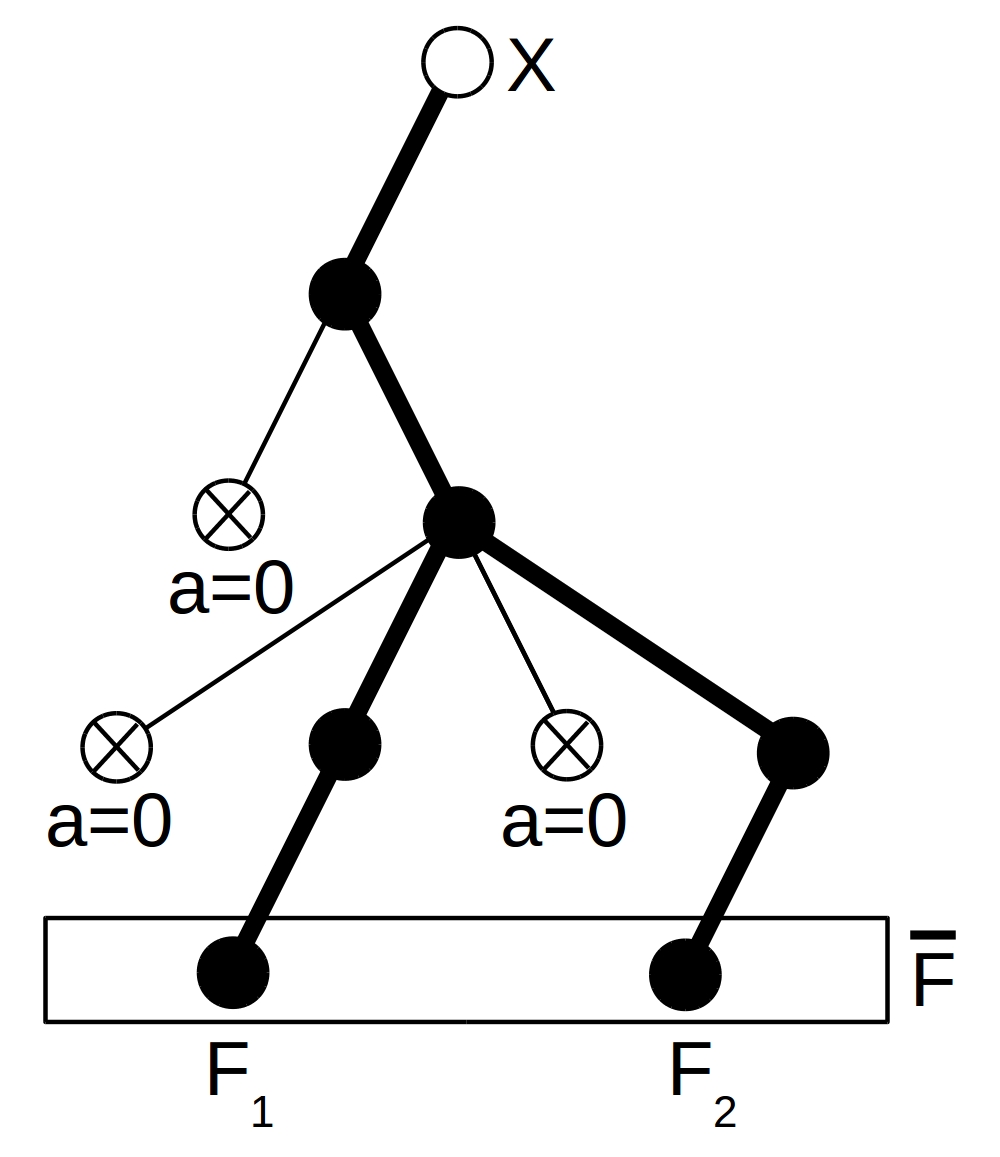}\\
\vspace{-0.15in}
\caption{The family of paths from the input pixel $X$ to output neurons $F_{1}$ and $F_{2}$ in the last set of feature maps $\bar{F}$. Crossed nodes correspond to neurons with zero activation. Paths containing these nodes do not contribute to $\bar{F}$.}
\label{fig:fig5}
\vspace{-0.05in}
\end{wrapfigure}
\begin{proof}
Note that averaging over different feature maps in the fixed convolutional layer with $f_{i}$ feature maps that is conducted by \textit{VisualBackProp} is equivalent to
summing over paths with edges in different feature maps of that layer (up to a multiplicative constant $\frac{1}{f_{i}}$.) Furthermore, adding contributions from different patches covering a fixed pixel $X$ is equivalent to summing over all paths that go through $X$. 
Thus computed expression is proportional to the union of products of activations (since in the algorithm after resizing the masks are point-wise multiplied) along paths from the given pixel $X$ in the input image to all the neurons in the last set of feature maps.
That leads to the proposed formula. Note that the formula automatically gets rid of all the paths that lead to a neuron with null activation since these paths do not contribute to activations in the last set of feature maps (see: Figure \ref{fig:fig5}). 
\end{proof}

\newpage
\section{Additional experiments}

\begin{table}[htp!]
\vspace{-0.15in}
\begin{center}
\setlength{\tabcolsep}{3.5pt}
\begin{tabular}{c|c|c|cc}
\cline{2-3}
  & {\textit{NetSVF}}  & {\textit{NetHVF}} &  & \\
\hline
\multicolumn{1}{|c|}{Layer}  & Layer  & Layer & \multicolumn{1}{|c|}{Filter}  & \multicolumn{1}{|c|}{Stride}\\
\multicolumn{1}{|c|}{}  & output size  & output size &  \multicolumn{1}{|c|}{size}  & \multicolumn{1}{|c|}{size} \\
\hline
\hline
\multicolumn{1}{|c|}{\textbf{conv}} &   $32\times 123\times 638$ &   $32\times 123\times 349$ & \multicolumn{1}{|c|}{$3\times 3$} & \multicolumn{1}{|c|}{$1\times 1$}        \\
\hline
\multicolumn{1}{|c|}{\textbf{conv}} &   $32\times 61\times 318$  &   $32\times 61\times 173$  & \multicolumn{1}{|c|}{$3\times 3$} & \multicolumn{1}{|c|}{$2\times 2$}        \\
\hline
\multicolumn{1}{|c|}{\textbf{conv}} &   $48\times 59 \times 316$  &   $48\times 59\times 171$  & \multicolumn{1}{|c|}{$3\times 3$} & \multicolumn{1}{|c|}{$1\times 1$}        \\
\hline
\multicolumn{1}{|c|}{\textbf{conv}} &   $48\times 29\times 157$  &   $48\times 29\times 85$   & \multicolumn{1}{|c|}{$3\times 3$} & \multicolumn{1}{|c|}{$2\times 2$}        \\
\hline
\multicolumn{1}{|c|}{\textbf{conv}} &   $64\times 27\times 155$  &   $64\times 27\times 83$   & \multicolumn{1}{|c|}{$3\times 3$} & \multicolumn{1}{|c|}{$1\times 1$}        \\
\hline
\multicolumn{1}{|c|}{\textbf{conv}} &   $64\times 13 \times 77$   &   $64\times 13\times 41$   & \multicolumn{1}{|c|}{$3\times 3$} & \multicolumn{1}{|c|}{$2\times 2$}        \\
\hline
\multicolumn{1}{|c|}{\textbf{conv}} &   $96\times 11\times 75$   &   $96\times 11\times 39$   & \multicolumn{1}{|c|}{$3\times 3$} & \multicolumn{1}{|c|}{$1\times 1$}        \\
\hline
\multicolumn{1}{|c|}{\textbf{conv}} &   $96\times 5\times 37$    &   $96\times 5\times 19$    & \multicolumn{1}{|c|}{$3\times 3$} & \multicolumn{1}{|c|}{$2\times 2$}        \\
\hline
\multicolumn{1}{|c|}{\textbf{conv}} &  $128\times 3\times 35$    &  $128\times 3\times 17$    & \multicolumn{1}{|c|}{$3\times 3$} & \multicolumn{1}{|c|}{$1\times 1$}        \\
\hline
\multicolumn{1}{|c|}{\textbf{conv}} &  $128\times 1\times 17$    &  $128\times 1\times 8$     & \multicolumn{1}{|c|}{$3\times 3$} & \multicolumn{1}{|c|}{$2\times 2$}        \\
\hline
\multicolumn{1}{|c|}{\textbf{FC}}   & 1024         & 1024         &   \multicolumn{1}{|c|}{-} &   \multicolumn{1}{|c|}{-}        \\
\hline 
\multicolumn{1}{|c|}{\textbf{FC}}   &  512         &  512         &  \multicolumn{1}{|c|}{-} &   \multicolumn{1}{|c|}{-}        \\
\hline
\multicolumn{1}{|c|}{\textbf{FC}}   &    1         &    1         &  \multicolumn{1}{|c|}{-} &   \multicolumn{1}{|c|}{-}       \\
\hline
\end{tabular}
\end{center}
\caption{Architectures of \textit{NetSVF} and \textit{NetHVF}. Each layer except for the last fully-connected layer is followed by a ReLU. Each convolution layer is preceded by a batch normalization layer. Let $n$ be the number of feature maps, $h$ be the height and $w$ be the width. For convolutional layers, layer output size is $n \times h \times w$. Filter size and stride are given as $h \times w$.} 
\label{tab:nnarch}
\vspace{-0.03in}
\end{table}

\begin{figure}[htp!]
\vspace{-0.25in}
\centering
\includegraphics[width=0.49\columnwidth]{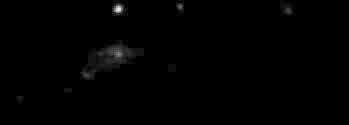} 
\includegraphics[width=0.49\columnwidth]{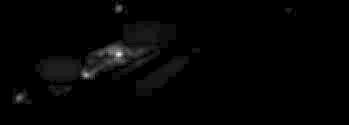} \\ 
\vspace{0.005in}
\includegraphics[width=0.49\columnwidth]{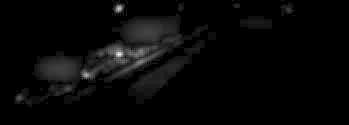} 
\includegraphics[width=0.49\columnwidth]{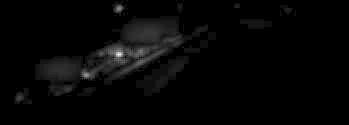} \\ 
\vspace{0.005in}
\includegraphics[width=0.49\columnwidth]{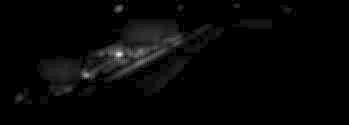} 
\includegraphics[width=0.49\columnwidth]{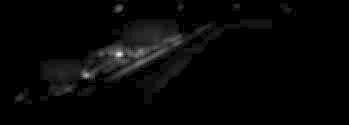} \\ 
\vspace{0.005in}
\includegraphics[width=0.98\columnwidth]{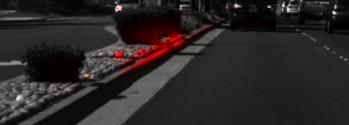} 
\caption{Visualizing the training of the network \textit{NetHVF} with the VisualBackProp method: the mask after 0.1, 0.5, 1, 4, 16, and 32 epochs. Over time the network learns to detect more relevant cues for the task from the training data. The last picture in each set shows the input image with the mask overlaid in red.} 
\label{exp_f6a2}
\end{figure}

\begin{figure}[!htp]
\centering
\includegraphics[width=0.495\columnwidth]{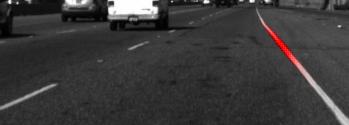}
\includegraphics[width=0.495\columnwidth]{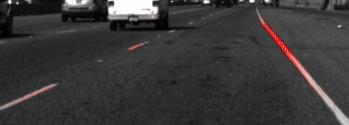} \\
\vspace{0.02in}
\includegraphics[width=0.495\columnwidth]{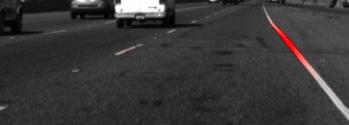}
\includegraphics[width=0.495\columnwidth]{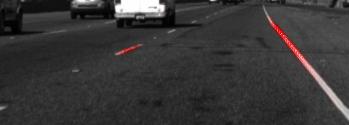}\\
\caption{Network \textit{NetHVF}. Left: VisualBackProp, right: LRP. Top and bottom of each column: two consecutive frames with the corresponding masks overlaid in red. The errors are (from the top): $1.55$ and $-0.66$ degrees of SWA.} 
\label{exp_f3b}
\vspace{-0.05in}
\end{figure}

\begin{figure}[htp!]
\vspace{-0.12in}
\centering
\includegraphics[width=0.495\columnwidth]{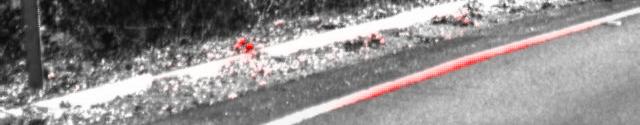}
\includegraphics[width=0.495\columnwidth]{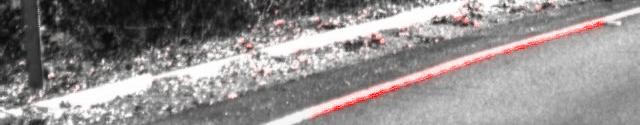} \\
\vspace{0.02in}
\includegraphics[width=0.495\columnwidth]{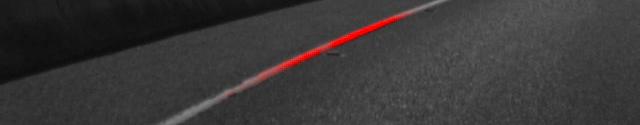}
\includegraphics[width=0.495\columnwidth]{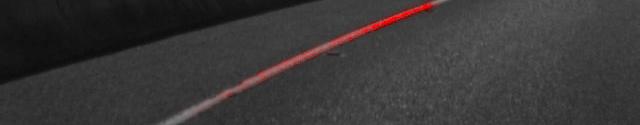}\\
\caption{Network \textit{NetSVF}. Left: VisualBackProp, right: LRP. Input test image with the mask overlaid in red. The errors are (from the top): $0.13$ and $-3.91$ degrees of SWA.} 
\label{exp_f4a}
\vspace{-0.05in}
\end{figure}

\begin{figure}[!htp]
\vspace{-0.12in}
\centering
\includegraphics[width=0.495\columnwidth]{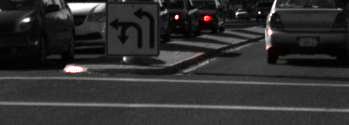}
\includegraphics[width=0.495\columnwidth]{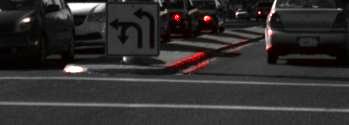} \\
\vspace{0.02in}
\includegraphics[width=0.495\columnwidth]{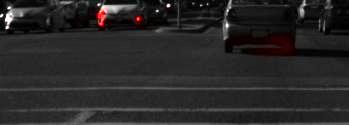}
\includegraphics[width=0.495\columnwidth]{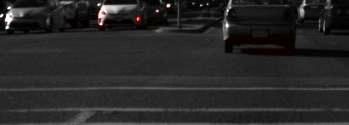} \\
\vspace{0.02in}
\includegraphics[width=0.495\columnwidth]{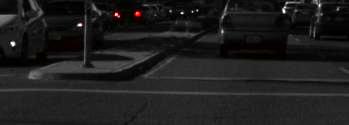}
\includegraphics[width=0.495\columnwidth]{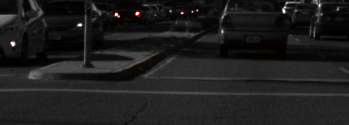}\\
\caption{Network \textit{NetHVF}. Left: VisualBackProp, right: LRP. Input test image with the mask overlaid in red. The errors are (from the top): $-7.72$, $-0.72$, and $-3.17$ degrees of SWA.} 
\label{exp_f5b}
\end{figure}

\begin{figure}[!htp]
\vspace{-0.12in}
\centering
\includegraphics[width=0.495\columnwidth]{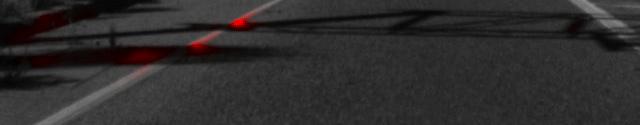}
\includegraphics[width=0.495\columnwidth]{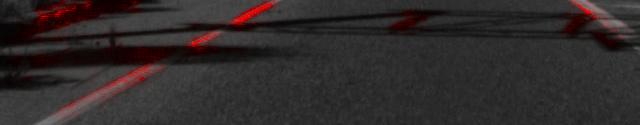}\\
\caption{Network \textit{NetSVF}. Left: VisualBackProp, right: LRP. Input test images with the corresponding masks overlaid in red. The error is 0.69 degrees of SWA.} 
\label{exp_f2a}
\vspace{-0.05in}
\end{figure}

\begin{figure}[!htp]
\vspace{-0.12in}
\centering
\includegraphics[width=0.495\columnwidth]{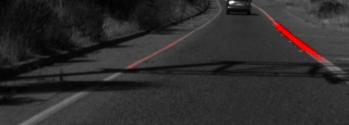}
\includegraphics[width=0.495\columnwidth]{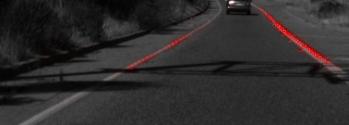}\\
\caption{Network \textit{NetHVF}. Left: VisualBackProp, right: LRP. Input test images with the corresponding masks overlaid in red. The error is -0.99 degrees of SWA.} 
\label{exp_f2b}
\end{figure}

Next we demonstrate the performance of VisualBackProp on the task of the classification of traffic signs. The experiments here were performed on the German Traffic Sign Detection Benchmark data set (\url{http://benchmark.ini.rub.de/?section=gtsdb&subsection=dataset}). We compare our method with LRP implementation as given in Equation 6 from~\cite{DBLP:journals/corr/SamekBMBM15} (similarly to the authors, we use $\epsilon = 100$). 

We train the neural network to classify the input image containing a traffic sign into one of 43 categories. The details of the architecture are described in Table~\ref{tab:nnarch2}. The network is trained with stochastic gradient descent (SGD) and the negative log likelihood (NLL) cost function for $100$ epochs.  

The German Traffic Sign Detection Benchmark data set that we use contains images of traffic signs of various sizes, where each image belongs to one out of 43 different categories. For the purpose of this paper, we scale all images to the size of $125\times125$.

We apply VisualBackProp and LRP algorithm on trained network. The results are shown in Figure~\ref{signs1}.

\begin{table}[htp!]
\vspace{-0.08in}
\begin{center}
\begin{tabular}{|c|c|c|c|}
\hline
{Layer}       & {Layer}      & {Filter}     & {Stride}\\
{}            & {output size}& {size}       & {size}\\
\hline
\hline
\textbf{conv} &   $16\times123\times123$ &  $3\times3$ & $1\times1$        \\
\hline
\textbf{conv} &   $16\times61\times61$   &  $3\times3$ & $2\times2$        \\
\hline
\textbf{conv} &   $24\times59\times59$   &  $3\times3$ & $1\times1$        \\
\hline
\textbf{conv} &   $24\times29\times29$   &  $3\times3$ & $2\times2$        \\
\hline
\textbf{conv} &   $32\times27\times27$   &  $3\times3$ & $1\times1$        \\
\hline
\textbf{conv} &   $32\times13\times13$   &  $3\times3$ & $2\times2$        \\
\hline 
\textbf{conv} &   $48\times11\times11$   &  $3\times3$ & $1\times1$        \\
\hline 
\textbf{conv} &   $48\times5\times5$     &  $3\times3$ & $2\times2$        \\
\hline
\textbf{FC}   &    64                    &   -         &   -        \\
\hline
\textbf{FC}   &    43                    &   -         &   -        \\
\hline
\end{tabular}
\end{center}
\caption{Architecture of the network used for sign classification. Each layer except for the last fully-connected layer is followed by a ReLU. The last fully-connected layer is followed by a LogSoftMax. Each convolution layer is preceded by a batch normalization layer.} 
\label{tab:nnarch2}
\vspace{-0.05in}
\end{table}

\begin{figure}[htp!]
\centering
\includegraphics[width=0.24\columnwidth]{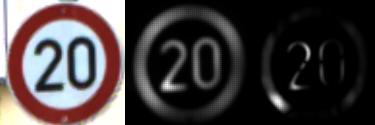} 
\includegraphics[width=0.24\columnwidth]{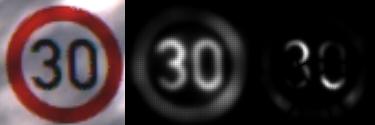}
\includegraphics[width=0.24\columnwidth]{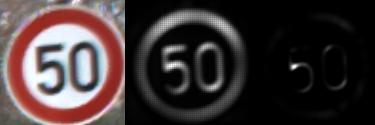} 
\includegraphics[width=0.24\columnwidth]{out175.jpg}\\
\vspace{0.02in}
\includegraphics[width=0.24\columnwidth]{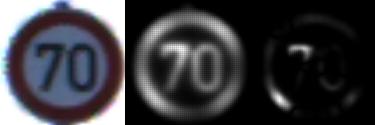} 
\includegraphics[width=0.24\columnwidth]{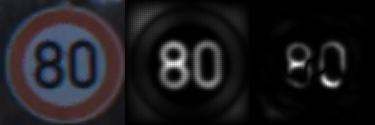}
\includegraphics[width=0.24\columnwidth]{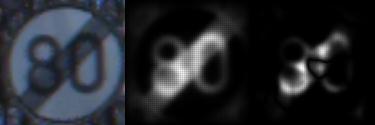} 
\includegraphics[width=0.24\columnwidth]{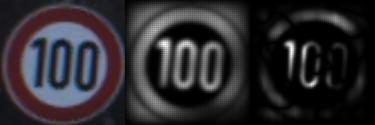}\\
\vspace{0.02in}
\includegraphics[width=0.24\columnwidth]{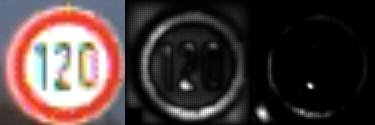} 
\includegraphics[width=0.24\columnwidth]{out433.jpg}
\includegraphics[width=0.24\columnwidth]{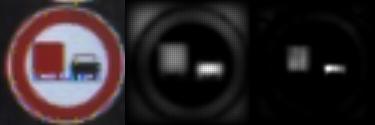} 
\includegraphics[width=0.24\columnwidth]{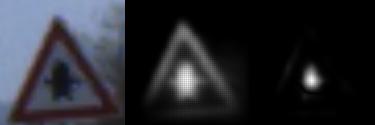}\\
\vspace{0.02in}
\includegraphics[width=0.24\columnwidth]{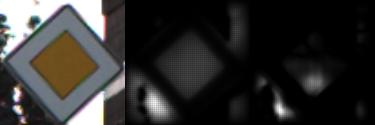} 
\includegraphics[width=0.24\columnwidth]{out678.jpg}
\includegraphics[width=0.24\columnwidth]{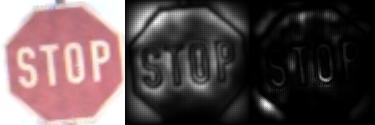} 
\includegraphics[width=0.24\columnwidth]{out792.jpg}\\
\vspace{0.02in}
\includegraphics[width=0.24\columnwidth]{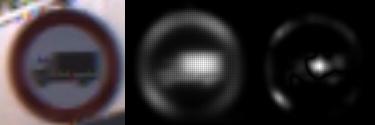} 
\includegraphics[width=0.24\columnwidth]{out819.jpg}
\includegraphics[width=0.24\columnwidth]{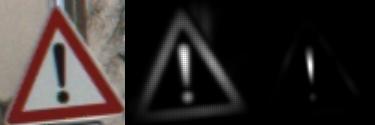} 
\includegraphics[width=0.24\columnwidth]{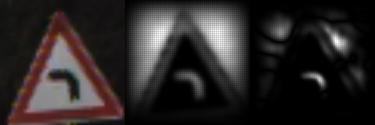}\\
\vspace{0.02in}
\includegraphics[width=0.24\columnwidth]{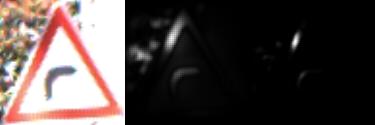} 
\includegraphics[width=0.24\columnwidth]{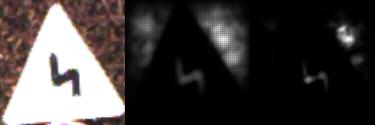}
\includegraphics[width=0.24\columnwidth]{out899.jpg} 
\includegraphics[width=0.24\columnwidth]{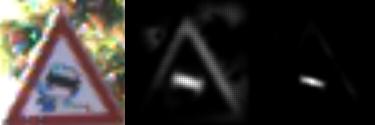}\\
\vspace{0.02in}
\includegraphics[width=0.24\columnwidth]{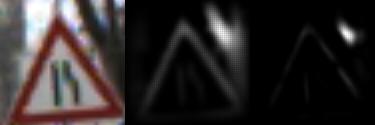} 
\includegraphics[width=0.24\columnwidth]{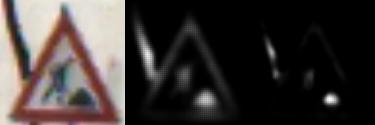}
\includegraphics[width=0.24\columnwidth]{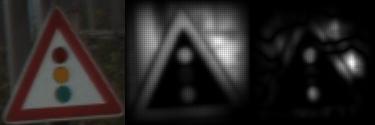} 
\includegraphics[width=0.24\columnwidth]{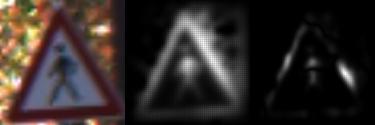}\\
\vspace{0.02in}
\includegraphics[width=0.24\columnwidth]{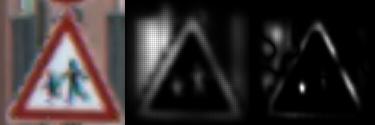} 
\includegraphics[width=0.24\columnwidth]{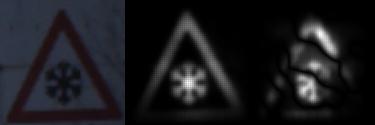}
\includegraphics[width=0.24\columnwidth]{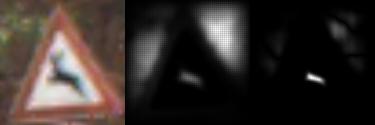} 
\includegraphics[width=0.24\columnwidth]{out1025.jpg}\\
\vspace{0.02in}
\includegraphics[width=0.24\columnwidth]{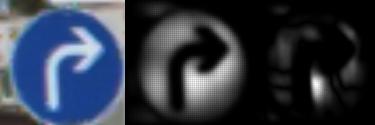} 
\includegraphics[width=0.24\columnwidth]{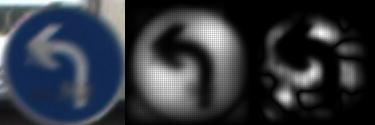}
\includegraphics[width=0.24\columnwidth]{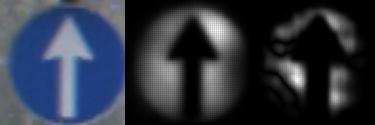} 
\includegraphics[width=0.24\columnwidth]{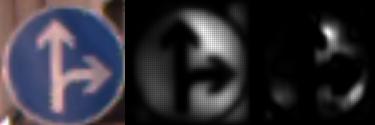}\\
\vspace{0.02in}
\includegraphics[width=0.24\columnwidth]{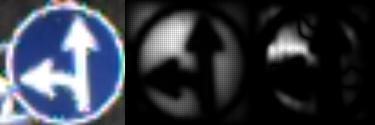} 
\includegraphics[width=0.24\columnwidth]{out1096.jpg}
\includegraphics[width=0.24\columnwidth]{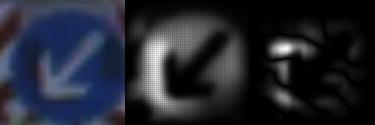} 
\includegraphics[width=0.24\columnwidth]{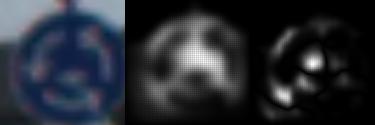}\\
\vspace{0.02in}
\includegraphics[width=0.24\columnwidth]{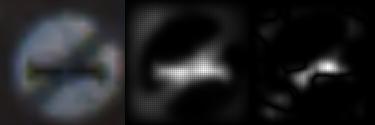} 
\includegraphics[width=0.24\columnwidth]{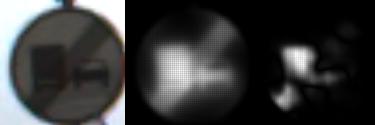}\\
\vspace{-0.05in}
\caption{Experiment on German Traffic Sign Detection Benchmark data set. Sets of input images with visualization masks arranged in four columns. Each set consist of an input image (left), a visualization mask generated by VisualBackProp (center), and a visualization mask generated by LRP (right).}
\label{signs1}
\vspace{-0.1in}
\end{figure}

\newpage
\begin{figure}[!htp]
\centering
\includegraphics[width=0.495\columnwidth]{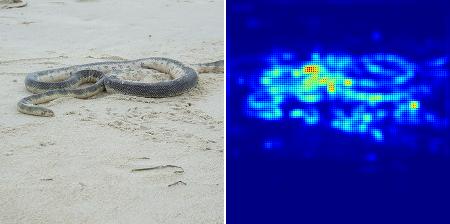} 
\includegraphics[width=0.495\columnwidth]{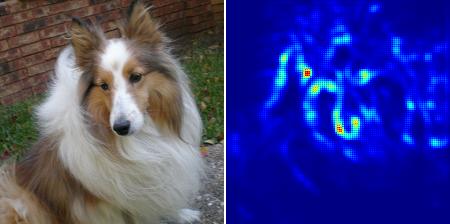}\\
\vspace{0.02in}
\includegraphics[width=0.495\columnwidth]{img3.jpeg} 
\includegraphics[width=0.495\columnwidth]{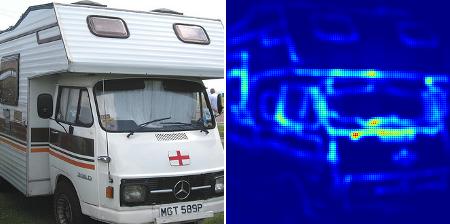}\\
\vspace{0.02in}
\includegraphics[width=0.495\columnwidth]{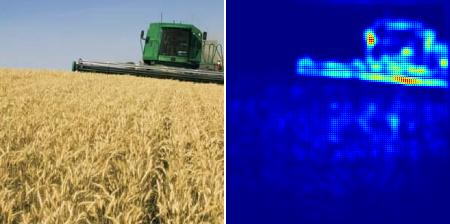} 
\includegraphics[width=0.495\columnwidth]{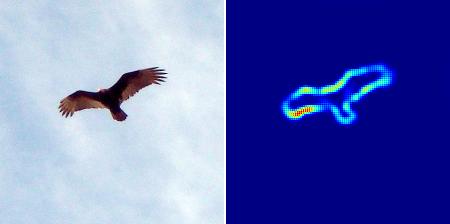}\\
\vspace{0.02in}
\includegraphics[width=0.495\columnwidth]{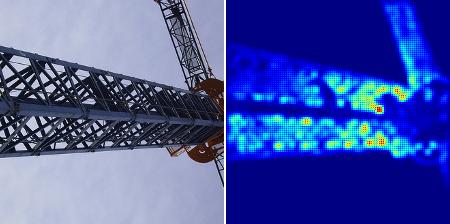} 
\includegraphics[width=0.495\columnwidth]{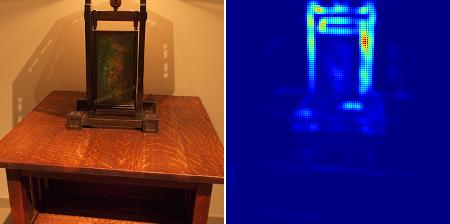}\\
\vspace{0.02in}
\includegraphics[width=0.495\columnwidth]{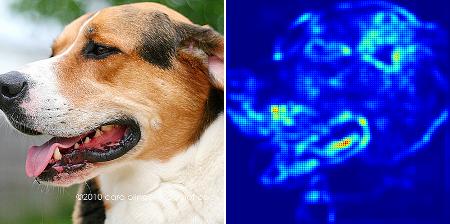} 
\includegraphics[width=0.495\columnwidth]{img10.jpeg}\\
\vspace{-0.05in}
\caption{Experiment on ImageNet data set. Sets of input images with visualization masks arranged in two columns. Each set consist of an input image (left), and a visualization mask generated by VisualBackProp (right).}
\label{resnet1}
\vspace{-0.1in}
\end{figure}

\newpage
\begin{figure}[!htp]
\centering
\includegraphics[width=0.495\columnwidth]{img11.jpeg} 
\includegraphics[width=0.495\columnwidth]{img12.jpeg}\\
\vspace{0.02in}
\includegraphics[width=0.495\columnwidth]{img13.jpeg} 
\includegraphics[width=0.495\columnwidth]{img14.jpeg}\\
\vspace{0.02in}
\includegraphics[width=0.495\columnwidth]{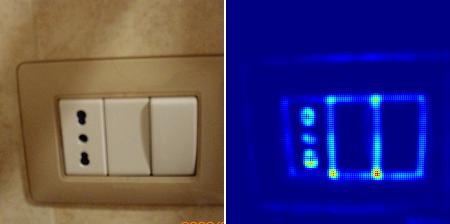} 
\includegraphics[width=0.495\columnwidth]{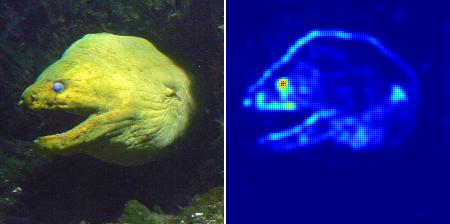}\\
\vspace{0.02in}
\includegraphics[width=0.495\columnwidth]{img17.jpeg} 
\includegraphics[width=0.495\columnwidth]{img18.jpeg}\\
\vspace{0.02in}
\includegraphics[width=0.495\columnwidth]{img19.jpeg} 
\includegraphics[width=0.495\columnwidth]{img20.jpeg}\\
\vspace{-0.05in}
\caption{Experiment on ImageNet data set. Sets of input images with visualization masks arranged in two columns. Each set consist of an input image (left), and a visualization mask generated by VisualBackProp (right).}
\label{resnet2}
\vspace{-0.1in}
\end{figure}

\end{document}